\documentclass{article}

\usepackage{arxiv}
\usepackage{doi}

\usepackage[utf8]{inputenc}
\usepackage[T1]{fontenc}

\usepackage{calc}
\usepackage[parfill]{parskip}
\usepackage{tipa}
\usepackage{longtable}
\usepackage{amssymb}
\usepackage{graphicx}
\usepackage{hyperref}
\hypersetup{
	colorlinks,
	linkcolor={red!75!black},
	citecolor={blue!50!black},
	urlcolor={blue!80!black}
}
\usepackage{textcomp}
\usepackage[
	natbib,
	style=authoryear,
	citestyle=authoryear,
	backend=biber]
	{biblatex}
\usepackage{citeposs}
\usepackage[linguistics]{forest}
\usepackage{booktabs}
\usepackage{tabularx}
\usepackage{multirow}
\usepackage{makecell}
\usepackage{wrapfig}

\newcommand{\hith}{\tipaLoweraccent[+.1ex]{\u{}}{h}}  %
\newcommand{\Hith}{\tipaLoweraccent[+.1ex]{\u{}}{H}} %
\newcommand{\shin}{\v{s}}
\newcommand{\SHIN}{\v{S}}
\newcommand{\heth}{\protect\hith}
\newcommand{\HETH}{\protect\Hith}
\newcommand{\eng}{\^g}
\newcommand{\ENG}{\^G}
\newcommand{\x}{\texttimes}

\usepackage{newunicodechar}
\newunicodechar{Š}{\SHIN}
\newunicodechar{š}{\shin}
\newunicodechar{Ḫ}{\HETH}
\newunicodechar{ḫ}{\heth}
\newunicodechar{ř}{\v{r}}
\newunicodechar{ğ}{\u{g}}
\newunicodechar{ū}{\=u} %

\newcommand{\htt}[1]{\emph{#1}}

\newcommand{\smr}[1]{\textsc{\MakeLowercase{#1}}}

\newcommand{\sign}[1]{\MakeUppercase{#1}}
\newcommand{\sub}[1]{\textsubscript{#1}}

\newcommand{\code}[1]{\texttt{#1}}

\newcommand{\stelzer}{kadaru}

\graphicspath{{./images/}}
\newlength{\normaldepth}
\makeatletter
\DeclareRobustCommand{\inlinesign}[1]{%
	\setlength{\normaldepth}{\depthof{p}}%
	\raisebox{-\normaldepth}{%
	\includegraphics[height={\f@size pt}]{#1}%
	}%
}
\makeatother

\onecolumn

\title{\textbf{A Recursive Encoding for Cuneiform Signs}} %
\author{
	\href{https://orcid.org/0009-0000-2865-3351}{\includegraphics[scale=0.06]{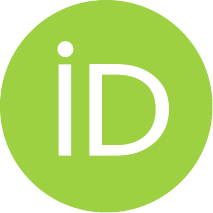}\hspace{1mm}Daniel M.~Stelzer} \\
	Department of Linguistics\\
	University of Illinois at Urbana-Champaign \\
	\texttt{stelzer3@illinois.edu}
}
\date{}

\hypersetup{
	pdftitle={A Recursive Encoding for Cuneiform Signs},
	pdfsubject={cs.CL},
	pdfauthor={Daniel M.~Stelzer},
	pdfkeywords={cuneiform, pedagogy, encoding, sign lists, sign lookup},
}

\begin{document}

\maketitle

\begin{abstract}
One of the most significant problems in cuneiform pedagogy is the process of looking up unknown signs, which often involves a tedious page-by-page search through a sign list. This paper proposes a new \emph{recursive encoding} for signs, which represents the arrangement of strokes in a way a computer can process. A series of new algorithms then offers students a new way to look up signs by any distinctive component, as well as providing new ways to render signs and tablets electronically.
\end{abstract}

\keywords{cuneiform \and pedagogy \and encoding \and sign lists \and sign lookup}

\section{Introduction}

Cuneiform is famous---or infamous---for several reasons. It is the oldest deciphered script in the world, the script used to document the oldest known epic poems and the oldest readable accounting records. And it is infamously difficult both to learn and to read. \citet[xxiv]{huehnergard} calls it ``very cumbersome'' and ``unquestionably the most difficult aspect of learning Akkadian'', relegating it to the later parts of his textbook. \citet[55]{cooper} claims that ``[n]either efficiency nor convenience played an important role in the development of Akkadian cuneiform'', and according to \citet[289]{worthington}, its orthographic features ``suggest that ancient sight-readers of cuneiform were expected to decipher a line a bit at a time---not to sweep their eyes across it as we do with our script.''

But this difficulty isn't just a property of the system itself. As \citet[2]{watkins} put it, ``the pedagogical tools are, in many cases, non-optimal''. Looking up unfamiliar words in a dictionary, physical or electronic, is standard practice when learning a new language. But cuneiform in particular involves at minimum hundreds of phonetic signs, and hundreds of logograms on top of those. Even after memorizing the basics, modern students will inevitably encounter signs they've never seen before---and on the level of individual signs, there's no alphabetical order or computerized search to help find them. While a learner of English has a standardized ordering to help their search (if they're looking for \emph{reverent}, it will be after \emph{revenge} and before \emph{revile}), a student of Akkadian or Sumerian has no such aid.

The traditional solution for this problem is a sign list, such as \posscite{borger} \citetitle{borger} or \posscite{labat} \citetitle{labat}. These typically order the signs based on their strokes, counted from left to right, in the clean, unambiguous Neo-Assyrian form. But most cuneiform is not Neo-Assyrian, and damaged tablets are the rule rather than the exception. Tablets tend to be found in a minimum of ten distinct pieces \citep[8]{gordin2}, and signs with damage to the left side are common. In these cases, \citet{robson3} recommends checking other sign indices, and offers a few alternatives if those also fail:

\begin{quote}
You can make an educated guess at the value the sign ought to have, based on the signs immediately around it, and then look up that value in the relevant index of Borger or Labat. Then you can compare your sign with the entry in Labat's table or Borger's paleographic list. The PSL lists of homophones and compounds can often be useful aids in this type of search.

Or you can simply page through Labat's table, looking for the closest match to your sign in the relevant script. I have done this countless times. It doesn't seem very clever or efficient, but sometimes it is the only way to find what you are looking for.
\end{quote}

The aim of this project is to find a better way. While learning cuneiform has always been a difficult and time-consuming task, we now have technologies the ancient Babylonians never dreamed of. Can we find a better way of looking up an unknown, potentially-damaged sign than scanning through Labat one page at a time?

Section~\ref{sec:bg} reviews previous work in this area, both with cuneiform and other logographic writing systems. Section~\ref{sec:encoding} then draws on this to propose a new encoding system for cuneiform, based on earlier work with Xixia. Section~\ref{sec:algo} describes several algorithms using this new system to ease the task of searching for an unknown logogram. Finally, section~\ref{sec:examples} provides examples of how this would be used in practice.

\section{Previous Work}
\label{sec:bg}

\subsection{Sign Lists}
\label{subsec:lists}

Unfamiliar logograms in cuneiform are not a new problem---one ancient letter describes a perplexed Kassite administrator receiving shipments of straw (the very common logogram \inlinesign{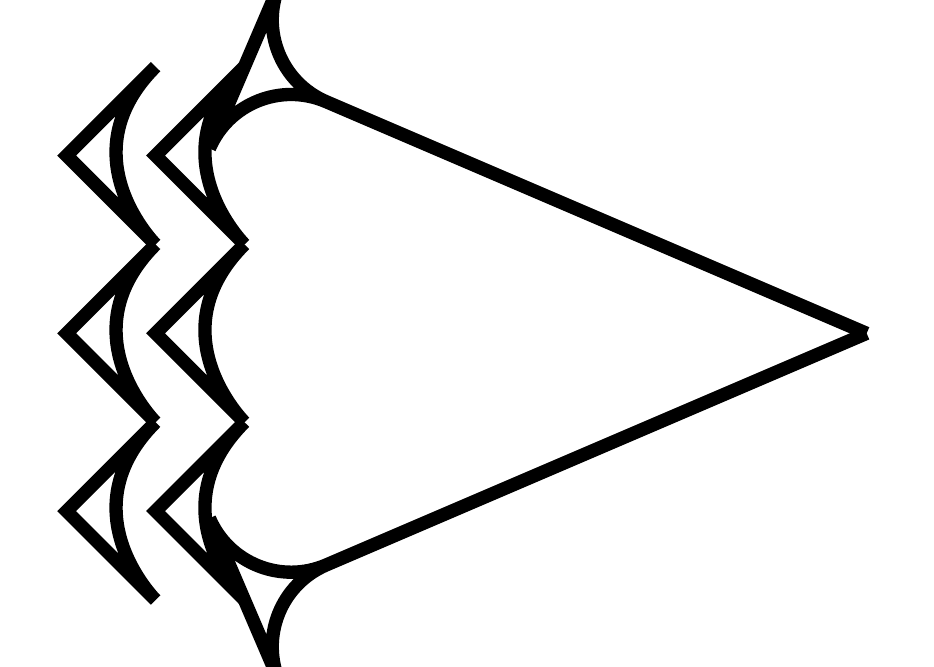} \smr{in}) when he'd ordered clay pots (the similar and much rarer \inlinesign{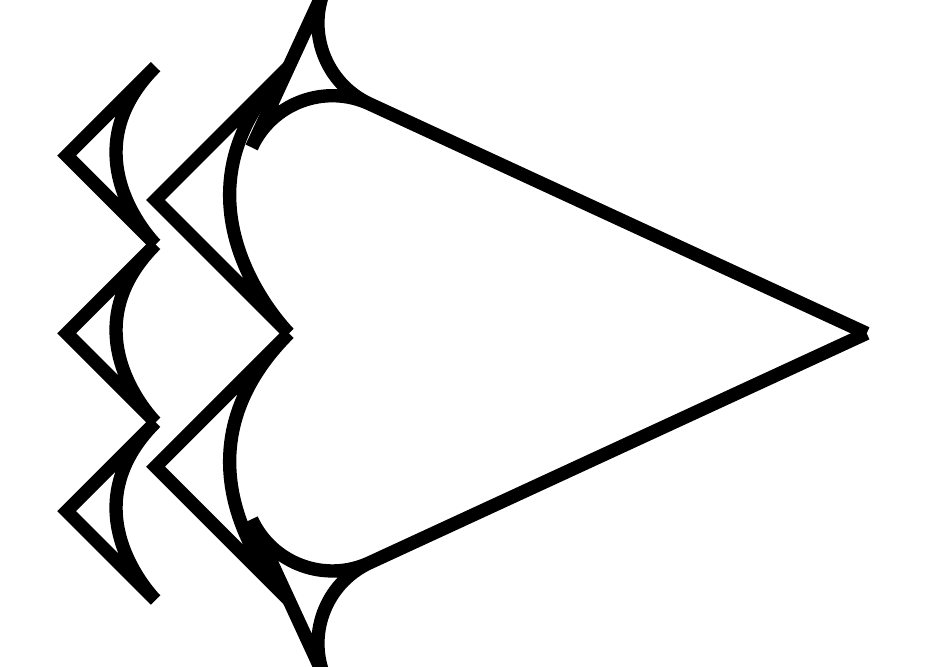} \smr{kan.ni}) \citep[viii, 142]{radau}. Ancient lexical lists giving names and meanings for cuneiform logograms are archaeologically common \citep{cooper}, as is evidence of students using them. Babylonian scribes referenced, copied out, and eventually memorized many of these sign lists to prepare for their job---as we can learn from their own complaints!\footnote{Taken from \citet[163]{sjöberg}.}

\begin{quote}
If you have learned the scribal art, you have recited all of it, the different lines [of the dictionary], chosen from the scribal art; the [names of] animals living in the steppe to the [names of] artisans, you have written; after that, you hate writing!
\end{quote}

However, it's unclear whether scribes would actually memorize every logogram, and texts describe scribes both reciting them and referencing physical tablets (from the same text as above: ``all the vocabulary of the scribes in the \emph{eduba} is in your hands''). \citet[164]{sjöberg} points out that, while scribes would brag about their memorization skills, the surviving lexical lists are often extremely long, and would be infeasible to memorize by rote. The aforementioned list of ``names of artisans'', \smr{lú} = \htt{ša}, consists of about a thousand lines \citep{taylor}, while the ``animals living in the steppe'' comes from \smr{ur\textsubscript{5}-ra} = \htt{ḫubullu}, with around 3,000 \citep{dcclt}. \citet[289]{worthington} goes further and suggests that reading fluently through a text simply did not happen in ancient Mesopotamia, with scribes frequently needing to pause to figure out an unclear or ambiguous sign.

In modern times, several authors have revived the ancient tradition of cuneiform sign lists, such as \citet{shl} for Sumerian, \citet{hzl} for Hittite, or \citet{borger} and \citet{labat} for Akkadian. These are generally extensive lists of signs with names and information on each one. While there's no universal order for cuneiform signs that could aid in searching, an informal standard has arisen: sorting starts at the left side of the sign, with \inlinesign{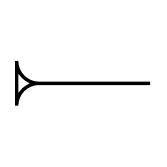} preceding \inlinesign{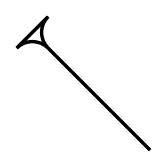} preceding \inlinesign{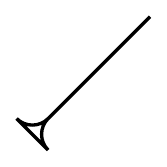} preceding \inlinesign{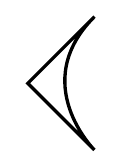} preceding \inlinesign{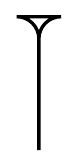} \parencites[1]{borger}[26]{labat}. \citeauthor{borger} set precedent basing his sorting on the Neo-Assyrian style, where the ordering of strokes tends to be fairly clear. But in his own words \citep[2]{borger2}:

\begin{quote}
Regrettably, it is practically impossible to arrange other versions of cuneiform writing (including the Ur III signs) by the shape of their signs in a consistent and unequivocal way.
\end{quote}

Indeed, while \posscite{labat} general Akkadian index and \posscite{hzl} Hittite sign list try to follow the same pattern, it's an imprecise measure at best. The ancients don't seem to have had any better system---the ordering of signs in standard lexical lists was quite arbitrary and generally seems to have come down to the whim of the compiler\footnote{See \citet[49--52]{cooper} for an extensive example.} \citep[48]{cooper}. Similar-looking and similar-sounding signs were generally grouped together, but without a broader sorting order as found in modern dictionaries \citep{crisostomo}.

More recently, electronic references such as \citet{signlist} and \citet{epsd} have arisen to make it easier to find information on particular signs. These are significantly more convenient than physical sign lists when looking up signs by name or reading, since they can take advantage of electronic searching. But they're no help when looking at an autograph or an actual tablet: the student needs to already know at least one reading to use these tools.

\subsection{Cuneiform Encodings}
\label{subsec:otherenc}

While sign lists and dictionaries remain the most popular cuneiform references, the present author is far from the first to notice the problem. According to \citet{gottstein}, these sign lists ``are very helpful for translation work, but are in most cases extremely impractical to handle''\footnote{\citet[127]{gottstein}: \emph{die zwar sehr hilfreich für die Übersetzungsarbeit, in der Handhabung jedoch grö\ss{}tenteils äu\ss{}erst unpraktisch sind}}---in particular, ``because of how comparatively time-consuming it is to find particular signs, especially in academic introductory classes''\footnote{\citet[127]{gottstein}: \emph{an dem vergleichsweise hohen Zeitaufwand, den das Auffinden bestimmter Zeichen [...] -- gerade auch im akademischen Anfängerunterricht -- mit sich bringt}}. In his words:

\begin{quote}
A precise system for classifying and, more importantly, looking up particular cuneiform signs within the framework of a analytical sign dictionary has long been among the desiderata of Ancient Near Eastern research, but so far has been neither realized nor tackled in any consistent way.\footnote{\citet[127--8]{gottstein}: \emph{Eine stringente Systematik zur Identifikation und vor allem Auffindung bestimmter Keilschriftzeichen im Rahmen eines analytischen Zeichenkompendiums gehört daher seit Langem zu den Desideraten der altorientalistischen Forschung, wurde bislang jedoch weder realisiert noch konsequent in Angriff genommen.}}
\end{quote}

A chemist by profession, Gottstein was inspired by molecular formulae. There's no obvious way to impose an alphabetical order on three-dimensional chemical structures, either, but reference works on chemistry have found a solution: molecules are indexed by the number of each type of atom they contain. A conventional order for the atoms ensures that these formulae can be sorted in a coherent way: the entry on nitrobenzene would be listed under C\sub{6}H\sub{5}NO\sub{2}, after quinone (C\sub{6}H\sub{4}O\sub{2}) but before benzene (C\sub{6}H\sub{6})\footnote{Like with cuneiform signs, chemicals tend to have names that can be alphabetized cleanly---but this is little help to a student who doesn't know the name of a new substance.}.

A molecular formula isn't necessarily unique---C\sub{6}H\sub{5}NO\sub{2} can also describe nicotinic acid, or a handful of other chemicals. But there are generally few enough chemicals with a particular formula that a student can scan through them easily to find the one they need. This is the sort of system Gottstein hoped to extend to cuneiform.

Cuneiform across all times, places, and languages is generally analyzed as having five basic types of wedges: vertical, horizontal, downward diagonal, upward diagonal, and the ``Winkelhaken'' or ``hook'' \parencites[9]{vandenhout}[1]{borger}. These became the ``elements'' of Gottstein's encoding, with the first four labelled `a', `b', `c', and `d' respectively\footnote{Gottstein's ordering of the strokes differs from what's generally used in sign lists, where horizontals come first and verticals last.}. After some early experiments, he grouped Winkelhaken into the `c' category as well---these identifiers weren't meant to be entirely unique, and the distinction between downward diagonals and Winkelhaken is not always obvious on an actual tablet. Still, some users of this system have extended it with a `w' category to capture that difference\footnote{This extension appears in \citet{paleocodage}, but that may not be its first usage. Wikidata terms it ``extended Gottstein encoding'', as seen at \url{https://www.wikidata.org/wiki/Q119228805}, where one particular sign variant is classified as \texttt{a13b5c1w2}.}.

\begin{figure}[h]
\centering
\includegraphics[width=0.67\linewidth]{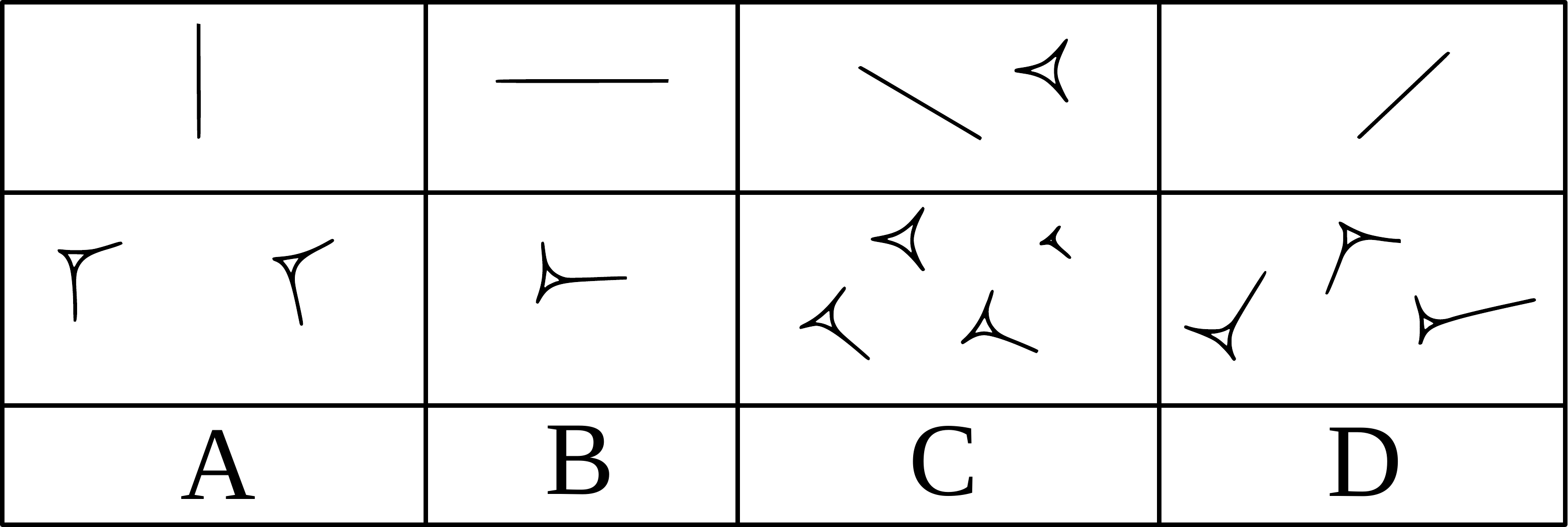}
\caption{A demonstration of the ``Gottstein system'', adapted from \citet[129]{gottstein} and \citet[ii131]{paleocodage}.}
\label{fig:gottstein1}
\end{figure}

\begin{wrapfigure}[14]{L}{0.4\textwidth}
\includegraphics[width=0.38\textwidth]{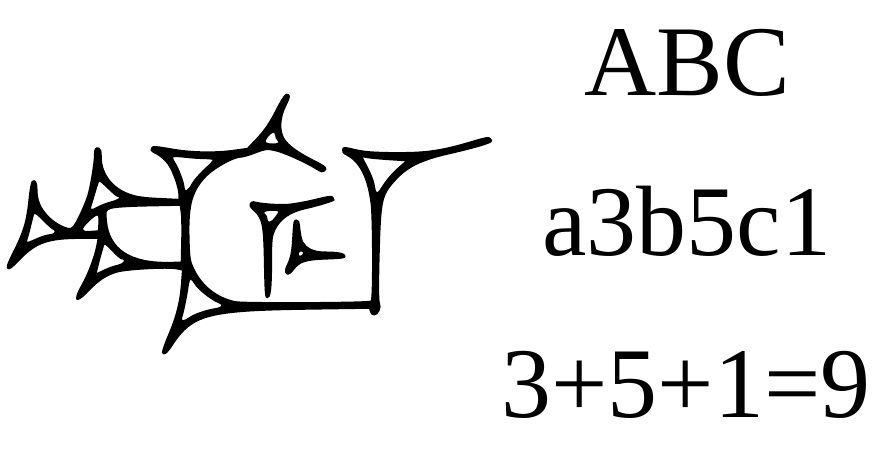}
\caption{Gottstein's analysis of the sign \smr{eme} `tongue', adapted from \citet[129]{gottstein} and \citet[ii131]{paleocodage}.}
\label{fig:gottstein2}
\end{wrapfigure}

Every sign can then be categorized in three ways: by the total number of strokes it contains (``category''), which \emph{types} of strokes it contains (``designation''), and the number of each type of stroke (``Gottstein code''). The sign \smr{EME} `tongue' in figure~\ref{fig:gottstein2} contains nine strokes total, of the `a', `b', and `c' species, giving it category 9, designation \texttt{ABC}, and Gottstein code \texttt{a3b5c1}.

\begin{figure}[h]
\centering
\includegraphics[width=0.75\linewidth]{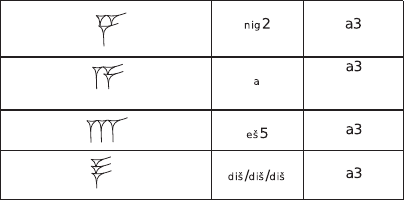}
\caption{An excerpt from \posscite[133]{gottstein} sign list, showing the signs with the Gottstein code \texttt{a3}.}
\label{fig:gottstein3}
\end{figure}

Gottstein proposed a sign list that would be organized first by category, then by Gottstein code, displaying each sign and variant that could possibly have that code: figure~\ref{fig:gottstein3} shows the section for category 3, code \texttt{a3}.

In the same paper, Gottstein describes the broad outlines of an experiment, reporting that ``every sign listed according to the `Gottstein System' could be found within seconds. Even non-specialists could find the signs they were looking for within a few moments.''\footnote{\citet[131]{gottstein}: \emph{jedes nach dem ``Gottstein-System'' gelistete Zeichen in Sekundenschnelle lokalisiert werden konnte. Sogar Fachfremde haben gesuchte Zeichen in wenigen Augenblicken gefunden.}} While several signs could have the same Gottstein code, this posed little difficulty: ``Experience shows that the additional effort only takes a second.''\footnote{\citet[131]{gottstein}: \emph{Der Mehraufwand beträgt erfahrungsgemä\ss{} nur eine Sekunde.}} This would seem to be a perfect solution to the problem, and indeed, it forms the basis of various electronic sign recognition systems like CuneiPainter\footnote{\url{https://situx.github.io/CuneiPainter/}}.

\begin{figure}[h]
\centering
\includegraphics[width=0.5\linewidth]{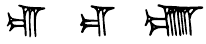}
\caption{Three variants of the sign \smr{U\sub{2}} `plant'. The leftmost is standard, but all three are attested in Hittite.}
\label{fig:u2var}
\end{figure}

However, cuneiform signs differ from chemicals in a few crucial respects. For the most part, adding or removing an atom from a molecule creates a different chemical completely: H\sub{2}O (water) behaves very differently from H\sub{2}O\sub{2} (hydrogen peroxide). But ancient scribes were less consistent. The sign \smr{U\sub{2}} `plant' is typically drawn with three vertical strokes, as shown in figure~\ref{fig:u2var}, but \citet[185]{hzl} report variations with as few as two and as many as seven!

Gottstein's solution is to list as many variants as possible, with entries for \sign{U\sub{2}} under \texttt{a2b2}, \texttt{a3b2}, and so on. More crucially, though, it's very common for a sign to be unclear or damaged. This is something Gottstein's system fundamentally cannot handle---if damage obliterated the only diagonal wedge in a sign, for example, its category, designation, and code would all be wrong!

\begin{wrapfigure}[12]{R}{0.25\textwidth}
\includegraphics[width=0.23\textwidth]{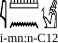}
\caption{The name of the god Amon, encoded in the MdC system.}
\label{fig:mdc}
\end{wrapfigure}

A different approach was taken by \citet{paleocodage}. In the study of Egyptian hieroglyphs, the \emph{Manuel de Codage} \citep{mdc} is a well-established standard for representing the layout of a text; it's intended to specify where each sign should be placed relative to the signs around it, making it easy to encode complicated arrangements of glyphs. For example, the MdC code in figure~\ref{fig:mdc} indicates that the game board (`mn') is on top of the water (`n') and to the right of the reed leaf (`i')\footnote{While it's not relevant here, the signs within the arrangement are named either by phonetic value or Gardiner code. See section~\ref{subsec:logograms}.}.

\posscite{paleocodage} system, ``PaleoCodage'', aims to encode cuneiform signs in the same way: by specifying the position of each stroke relative to its surroundings. The \emph{Manuel} defines only three operators\footnote{\texttt{-} separates groups of signs, \texttt{:} stacks signs vertically within a group, and \texttt{*} juxtaposes signs horizontally within a group. Most implementations add a fourth operator, \texttt{\&}, for special cases (``ligatures'') that don't follow any general pattern.}, which is sufficient for most hieroglyphic texts. But as the name suggests, PaleoCodage is intended for palaeography---for analyzing the fine details of handwriting and ductus that go beyond telling one sign from another. For this purpose, knowing that one stroke is to the right of another isn't enough. The exact distance there could be crucially important!

As a result, \citet[ii133--ii135]{paleocodage} extends Gottstein's four stroke types with the Winkelhaken, two types of reversed diagonal wedges, and two types of ``seal wedges'' used in archaic number signs, all of which can be modified in three different ways. These can be combined via a staggering array of operators, allowing a palaeographer to encode the size, angle, and position of each stroke with perfect accuracy. Three different ``to the right of'' operators encode subtle differences of spacing, or can be combined for further detail---and if this still proves insufficient, a ``factor operator'' can be used to adjust them by as little as one percent.

When discussing specific variants of signs, as \citet[ii138]{paleocodage} puts it, ``a very fine-granular modeling is often needed to depict the changes distinguishing the new sign variant from the other standard sign form''. PaleoCodage is thus tuned to specify enough detail to distinguish one scribe's particular handwriting from another. But this abundance of detail poses a difficulty for our purposes. \citet[ii140]{paleocodage} proposes a way to look up similar characters by comparing their PaleoCodage encodings---but by their string similarity metrics, a small horizontal stroke (\texttt{sb}) is no more similar to a large horizontal (\texttt{B}) than it is to a small vertical (\texttt{sa})!

In order for PaleoCodage to express this level of precision, its operators don't quite encode \emph{relationships} (as in the MdC system) between the strokes. Instead, each operator represents a change in state for the parsing automaton \citep[ii140]{paleocodage}. This means that the operators between particular wedges don't necessarily have anything to do with those wedges themselves---two vertical strokes next to each other might be encoded as \texttt{a-a} or \texttt{a\_a}, but the difference has nothing to do with those wedges themselves. Instead, it indicates whether any horizontal strokes crossing those wedges should cross both, or only one! In short, the same features that make PaleoCodage useful for palaeography also make it generally unsuitable for our purposes.

\subsection{Machine Learning}

In recent years, the problem of cuneiform transcription has attracted more attention from computer scientists, and machine learning algorithms have been applied to a variety of different aspects. Machine learning has been extremely effective at recognizing Han logograms, for example, even with only ``off-the-shelf'' algorithms \citep{kamate}. It would seem reasonable to apply these methods to cuneiform in the same way.

Unfortunately, the most effective algorithms for recognizing Han characters (as described by \citet{liuxin} and \citet{liuchenglin}, among others) tend to rely on specific details of how those characters are written, such as stroke trajectory. In theory, the concepts behind these algorithms could be adapted to cuneiform strokes. At present, however, the cutting-edge algorithms used in Han lookup tools (such as Pleco) cannot be easily applied to cuneiform. To make them work, some fundamental aspects would need to be redesigned from the ground up.

Unicode attempts to provide a codepoint for each cuneiform sign\footnote{Though it has its flaws: \citet[2]{borger2} is quite scathing in his criticism of the implementation.}, and most attempts at cuneiform machine learning, such as \citet{doostmohammadi} and \citet{gordin}, take Unicode-encoded text as their starting point. While remarkable progress has been made in tasks like identifying the language of a text and choosing appropriate readings for each sign, these projects require the signs to have already been transcribed and identified---which can often be the most difficult and time-consuming part. Research in this direction won't help with the actual lookup and identification of logograms.

Other projects, like \citet{dencker}, attempt to go all the way from tablet photographs to a full transliteration. These projects have also achieved remarkable successes from relatively little training material. But they often suffer from trying to do too much at once---the system has to learn the entire model at the same time, all the way from recognizing shadows in a photograph to picking out the right reading for a sign. This means that what the system learns about recognizing wedges in Neo-Assyrian cuneiform, for example, can't be easily generalized to another dialect like Old Akkadian, even though the fundamental principles are the same. This becomes a serious issue when there's very limited data for a particular dialect.

One remarkable outlier is \citet{kriege}, which builds on earlier work by \citet{fisseler}. \citeauthor{fisseler} aimed to solve a smaller problem: converting scans of tablets into readable autographs. They were remarkably successful at this, and \citeauthor{kriege} used the output of that model as the input to theirs, looking at relationships between wedges rather than raw images. By applying graph-based neural networks from \citet{fey} to this input, they were able to identify signs with remarkable accuracy.

However, their initial experiment used a very limited selection of signs written by modern scholars, with manual annotation for which wedges belonged to which signs. While their results are still extremely promising, and hint that specialized stroke-detection algorithms along the lines of \citet{liuxin} and \citet{liuchenglin} could be developed for cuneiform, it remains to be seen how well they will generalize to actual ancient tablets with hundreds of distinct signs and no clear marking of sign boundaries\footnote{A notable result in this direction is \citet{stötzner}, who attempt to automatically partition tablets into signs and recognize the location and direction of each wedge. An experiment in recognizing signs using the output of this model is currently in progress.}.

Based on this, it seems that there is currently no effective machine-learning solution to the problem of sign recognition. But note also that one of the most pressing difficulties in applying machine learning to cuneiform seems to be the lack of a good intermediate representation: a concise, useful way of indicating the wedges that make up a sign, which image-recognition models could convert scans and photographs into, and other models could separately convert to readings\footnote{\citet{ice} hoped that Unicode would form this intermediate representation, and as \citet{doostmohammadi} and \citet{gordin} showed, this is useful for separating out language detection and sign interpretation. But choosing the right sign is significantly more context-sensitive than \citeauthor{ice} expected, and it's not clear what advantages Unicode encoding has over Romanized sign names or index numbers---U+1227A is no less opaque than \htt{pa} or HZL 174 as a representation of \inlinesign{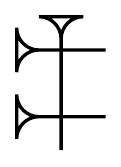}.}. This would effectively break the problem in half, like how modern machine learning models generally handle OCR (recognition of the graphemes making up written text) separately from the interpretation of that text. The new recursive encoding system proposed in this paper may prove useful for this purpose as well.

\subsection{Other Logographies}

\label{subsec:logograms}

For many languages, ancient and modern, learning the script is only a minor obstacle. Students of Greek generally have no issue learning its alphabet, for instance, and have no need for transliterations. Why is cuneiform any different?

The difference lies in the sheer number of signs. The Greek alphabet uses only 24 letters, while \citet{hzl} have documented 375 signs for Hittite, plus additional variations. Ten of those variations have distinct forms and meanings, raising the total to 385. For Akkadian, \posscite{labat} index numbers go up to 567, while \posscite{borger} reach 905.

Of the 385 signs used in Hittite, 230 of those are never used phonetically for Hittite words---they're exclusively used for logograms, punctuation, or foreign words\footnote{214 of those 230 are used as logograms, one (the ``Glossenkeil'') only as punctuation, and 15 for transcribing non-Hittite sounds. For example, Hurrian words sometimes contain ligatures of the sign \htt{wa} with various vowels, representing a labial fricative of some sort (/fa/, /fi/, etc) that Hittite lacked.}. So while the majority of Hittite text is written in phonograms, students are likely to come across a vast assortment of logograms in their studies---many of which they may never have memorized, or ever seen before. This is where a lookup system is essential.

Cuneiform, though, is not the only writing system to require hundreds of logograms. Various other scholars have run into this same problem when analyzing Han logograms and Egyptian hieroglyphs, among others, and have needed effective lookup methods long before the advent of machine learning.

\begin{figure}[h]
\centering
\includegraphics[width=0.8\linewidth]{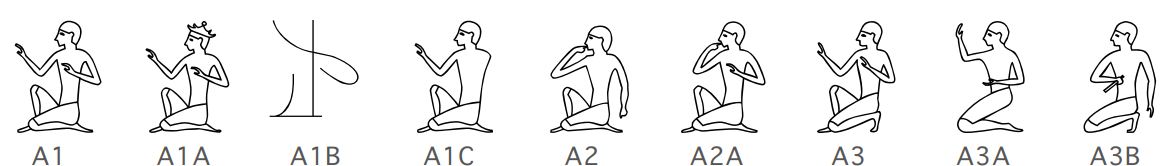}
\caption{A handful of Egyptian hieroglyphs classified with \citeauthor{gardiner}'s system.}
\label{fig:gardiner}
\end{figure}

For Egyptian hieroglyphs, the standard way to look up a sign originates with \citet{gardiner}\footnote{First published in complete form in \citet{gardiner2}, but widely popularized in his 1957 grammar.}. He created a catalogue of hieroglyphs based on the objects they represent, so that, for example, all signs depicting birds can be found under ``G'', and all signs depicting ships and parts of ships can be found under ``P''. This means that a student should only have to look through a few dozen signs to find the one they need, rather than several hundred. (See figure~\ref{fig:gardiner}.)

But Gardiner's system depends on students being able to tell easily what a sign depicts, which is sometimes non-trivial (it's not at all obvious to a modern-day student that a speckled circle represents a threshing floor while a lined circle represents an animal placenta\footnote{Or perhaps a woven mat, based on the color it's painted in certain inscriptions---experts disagree. Given how much the paint colors vary, it's likely that ancient scribes did too!}) and sometimes impossible (such as in hieratic writing, a much more stylized variation of hieroglyphic Egyptian). Certain signs could also reasonably fall into multiple categories, increasing the difficulty for the student. Would a sign of a vulture god and a cobra god together be found under G for ``birds'', I for ``reptiles and amphibians'', or C for ``deities''?\footnote{The answer is ``birds'', if you were wondering: sign G16, ``the Two Ladies''.} Cuneiform signs are much less representational than hieroglyphic ones, so this system is infeasible for our purposes.

\begin{figure}[h]
\centering
\includegraphics[width=0.5\linewidth]{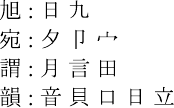}
\caption{A selection of Han characters and the radicals they contain, taken from \protect\citet[5]{breen}. The character can be looked up by one or more of these radicals. (From the top: `rising sun', `address', `origin', and `poetic meter'.)}
\label{fig:radicals}
\end{figure}

For the several thousand Han logograms, quite a lot of different lookup systems have been proposed over the years. The most famous and popular are radical-based systems \citep{breen}. Most Han logograms contain recognizable smaller parts, known as ``radicals''. An index of characters by their most distinctive radical, or (in newer electronic dictionaries) by \emph{all} radicals they contain, can then narrow down the search space tremendously \citep{breen}. (See figure~\ref{fig:radicals}.) Some go a step further, annotating the radicals for their exact positions within the sign, though the reliance on absolute positioning becomes a problem when signs may be obscured or damaged.

\begin{wrapfigure}[10]{L}{0.25\textwidth}
\includegraphics[width=0.23\textwidth]{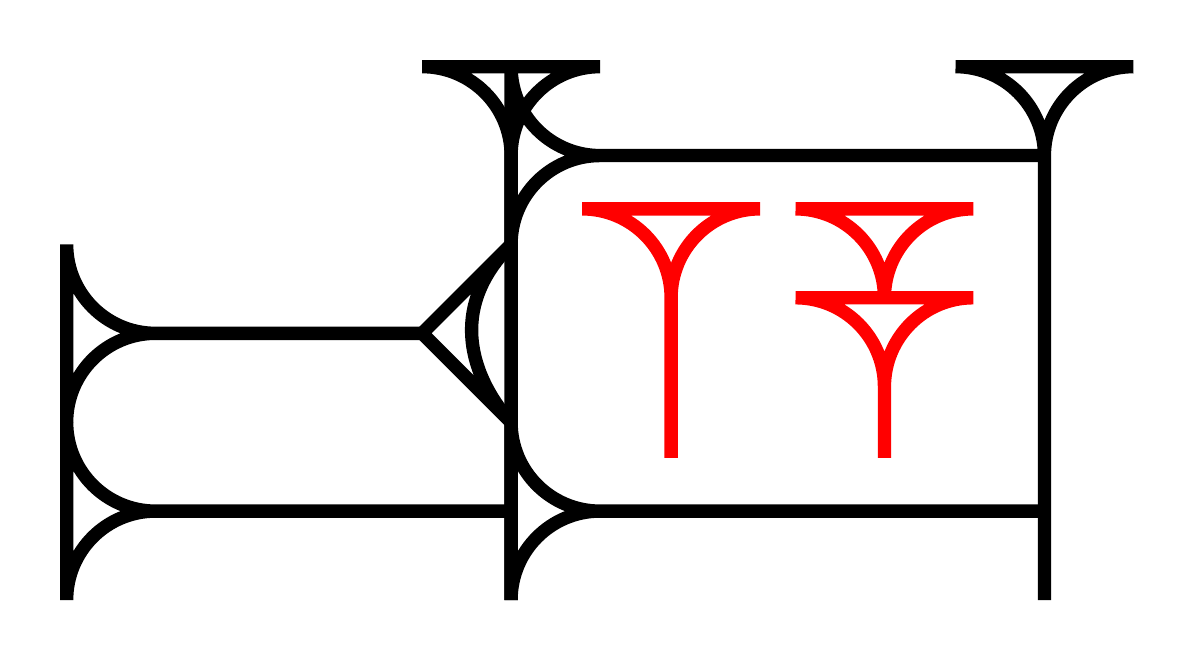}
\caption{\sign{KA} (black) enclosing \sign{A} (red).}
\label{fig:nag}
\end{wrapfigure}

To some extent, this sort of system has already been used for cuneiform. The naming rules for signs, as discussed in \citet{oracc} and \citet{robson2}, can indicate when a sign is composed of recognizable pieces: the logogram \smr{nag} `drink' (shown in figure~\ref{fig:nag}) is named \sign{KA\x{}A} (``\smr{ka} `mouth' enclosing \smr{a} `water'\,''), while the logogram \smr{ug} `tiger(?)' is officially named \sign{PIRI\ENG\&UT} (``\smr{piri\eng} `lion' on top of \htt{ut}'')\footnote{The \htt{ut} here hints at the pronunciation, distinguishing it from, say, \sign{PIRI\ENG\&ZA} for \smr{az} `bear'. Compare the use of phonetic complements in Han logograms.}. However, most cuneiform signs don't contain meaningful sub-units larger than single wedges, and signs that were once made of clearly distinct radicals may cease to be so over time: the logogram \smr{meš} `[plural]' was once a transparent compound \sign{ME+EŠ} (``\htt{me} up against \htt{eš}''), but became a single indivisible unit \sign{MEŠ} in later eras.

A related system indexes characters by how many strokes they contain; the KANJIDIC database that underlies popular tools like Jisho, for example, includes this type of indexing \citep{kanjidic}. However, as mentioned in section~\ref{subsec:otherenc}, the number of wedges in a cuneiform sign is much less consistent than the number of strokes in a Han logogram. Scribes would very frequently leave a wedge off, or include an additional one by mistake. This makes stroke-number systems generally unsuitable for cuneiform.

\begin{wrapfigure}[22]{R}{0.25\textwidth}
\includegraphics[width=0.23\textwidth]{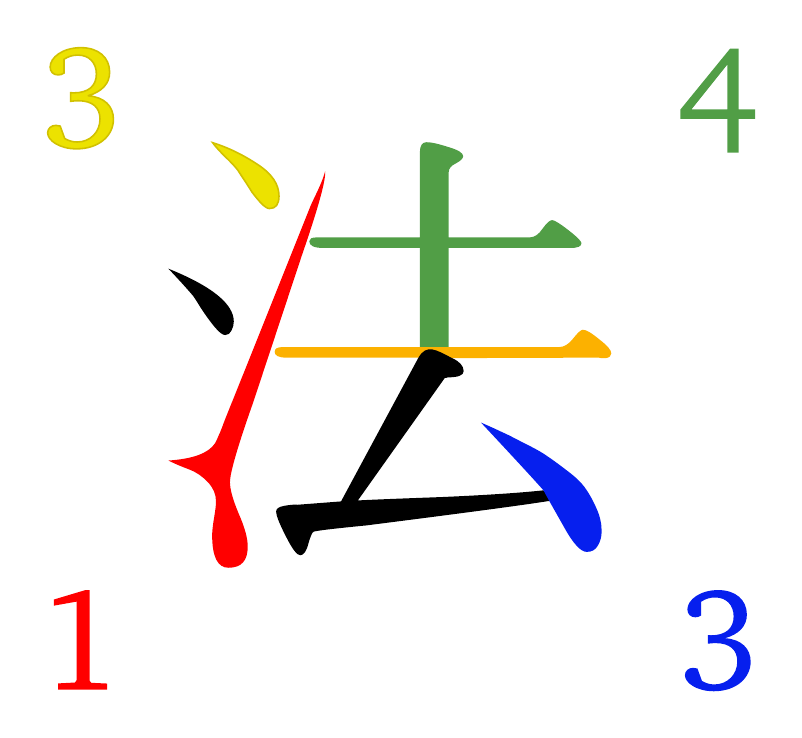}
\caption{An example of four-corners classification: this sign would be indexed as 3413 in a four-digit system, or 3413-1 in a five-digit system. Diagram by Oona Räisänen, taken from \url{https://commons.wikimedia.org/wiki/File:Four-corner_method.svg}.}
\label{fig:fourcorner}
\end{wrapfigure}

Another popular system for Han logograms is the ``four corners method'' \citep{fourcorner,downes}, which involves dividing all visual elements into ten broad categories, then listing which category each corner of the sign (and sometimes additionally the center) falls under. This gives each glyph a four- or five-digit number that can be used as an index, as seen in figure~\ref{fig:fourcorner}. A system broadly similar to this (using the leftmost edge) is already standard in cuneiform sign lists, as discussed above. It can be somewhat helpful; however, it still often leaves hundreds of glyphs to search through, damage to signs is common, and cuneiform scribes were often less careful than Han writers in keeping their stroke types distinct. When the leftmost stroke is used as the main index, uncertainty about that particular stroke---whether caused by scribal carelessness, damage to the tablet, or simply bad lighting in a photograph---can and does render the system unusable.

\begin{figure}[h]
\centering
\includegraphics[width=\linewidth]{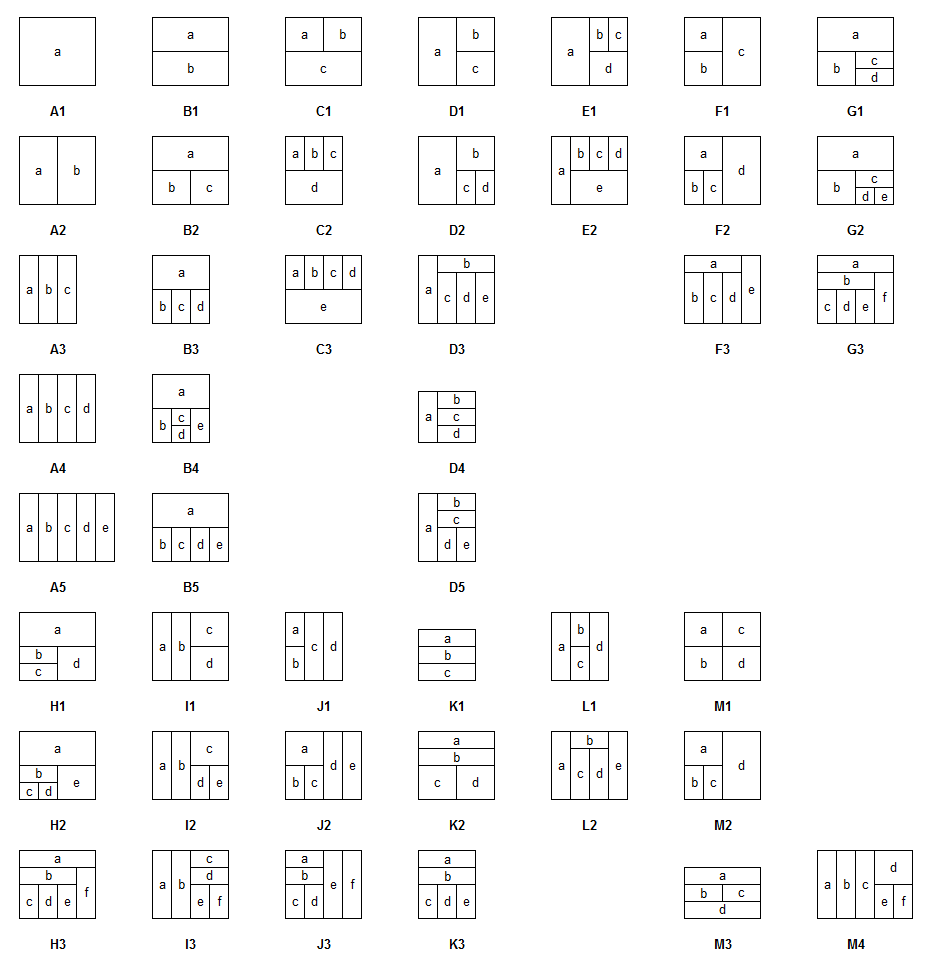}
\caption{The structures used in Nishida's index, from \citet[246]{nishida}, reproduced in \citet[13]{downes}.}
\label{fig:nishida}
\end{figure}

For Xixia/Tangut, an extinct logography possibly related to Han characters, a different sort of index was proposed by \citeauthor{nishida} in \citeyear{nishida}\footnote{The original source is \citet{nishida}, but as I don't speak Japanese, I'm relying on the synopsis in \citet[12--13]{downes}.}. Nishida's system divides all Xixia characters into 319 radicals, then indexes characters based on their structure---three radicals next to each other are structure A3, for example, while two radicals on top of two other radicals are structure M1 (see figure~\ref{fig:nishida}). Unfortunately, as discussed above, the dependence on radicals makes this system generally unsuitable for cuneiform.

\begin{figure}[h]
\centering
\includegraphics[width=0.67\linewidth]{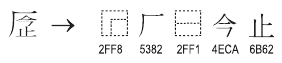}
\caption{An example of the IDS, from \citet[691]{unicode}.}
\label{fig:unicode}
\end{figure}

A related proposal was incorporated into Unicode in 1999, dubbed the ``Ideographic Description System'' and meant to apply to Han, Xixia, and other similar writing systems; the intent was to allow fonts to synthesize obscure characters that may not have dedicated glyphs. The IDS is similar to Nishida's index, enumerating a certain number of possible structures that radicals can fit into. But it goes one step further, allowing these structures to be nested recursively, as shown in figure~\ref{fig:unicode} \citep[689--692]{unicode}.

While this proposal has potential, the Unicode consortium warns against using it for anything beyond describing unencoded variants, and thus little time and effort has been devoted to it. At the time of writing, no software has been found that actually uses or even supports it. A similar system was proposed for cuneiform by \citet{ice2}, but was not accepted by the Unicode consortium\footnote{See \url{https://unicode.org/mail-arch/unicode-ml/y2004-m02/0012.html} for some discussion of this.}. \citet{wong} lay out another idea in the same vein, dubbed ``HanGlyph'', which uses 41 basic strokes and 12 operations to potentially describe \emph{any} Han character (even one not composed of recognizable radicals). But this proposal is also mostly theoretical, as no actual implementation seems to be available, or any evaluation of its usability.

\begin{figure}[h]
\centering
\includegraphics[width=0.75\linewidth]{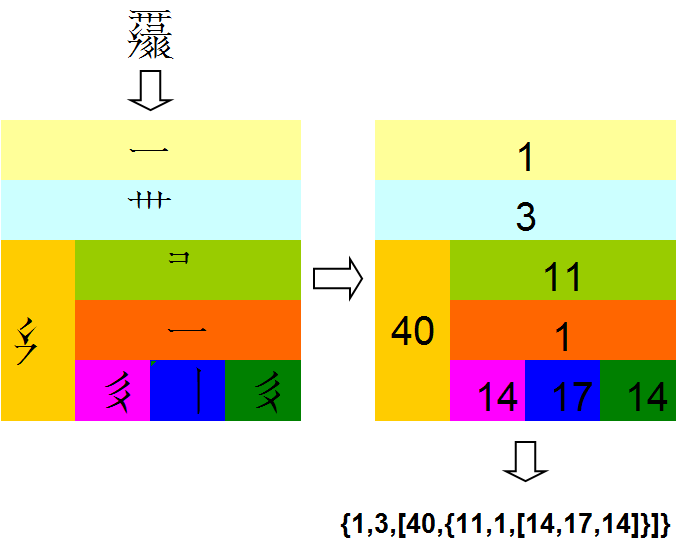}
\caption{An example of \citeauthor{downes}'s recursive index for Xixia, reproduced from \citet[15]{downes}.}
\label{fig:downes}
\end{figure}

Another recursive description system for both Xixia and Han was proposed by \citet{downes} and elaborated on in \citet{downes2}. Like the Unicode IDS, \citeauthor{downes}' system tries to describe the relationships between radicals in a recursive way. This system uses only three relationships, compared to the IDS's twelve: ``stacked horizontally'', ``stacked vertically'', and ``enclosed by''. These prove sufficient to describe virtually all Han and Xixia characters; the Xixia characters in fact can be described with only two relationships, since they never enclose one radical with another\footnote{\citet[13--14]{downes} relates a legend: this ensures that malevolent spirits can always escape and can't get trapped inside the character while it's being written.}. An example is given in figure~\ref{fig:downes}.

This system seems the most promising for our purposes\footnote{Though no studies have been found actually putting Downes' work into practice.}. Rather than \citeauthor{downes}'s 176 radicals for Xixia and 420 for Han, we can reduce characters all the way down to their most basic strokes. As mentioned in section~\ref{subsec:otherenc}, cuneiform across all times, places, and languages is generally analyzed as having five types of wedges---far fewer than \citeauthor{wong}'s 41. Empirically, only three types of compositions have proven sufficient to encode 99\% of the signs and variants in \citet{hzl}. Horizontal and vertical stacking are implemented as in \citet{downes}; the third form of composition is \emph{intersection} or \emph{superposition}, since this system goes down to the level of individual cuneiform wedges, while \citeauthor{downes} generally calls any intersecting strokes a new radical.

\section{Recursive Encoding}
\label{sec:encoding}

This new proposal---dubbed the ``kadaru'' system, from the Akkadian for ``subdivide''---is to represent a cuneiform glyph as a tree. The leaves of this tree are the five basic strokes, and the branches are the three basic ways of combining them (``compositions''): stacked horizontally, stacked vertically, or intersecting. Informal experiments suggest that \posscite{downes} bracket notation (\code{\{[vv][vv]\}}) is easier for students to read than \posscite{ice2} infix notation with precedence (\code{v+v\&v+v})\footnote{Infix notation with precedence is also used by the \emph{Manuel de Codage} itself, though \emph{not} by PaleoCodage, which has too many operators for precedence to be feasible.}, so this becomes the foundation of the syntax: strokes are indicated by lowercase letters\footnote{Different from the letters used by \citet{gottstein} and \citet{paleocodage}, for disappointingly mundane reasons---the foundations of this system were built before the authors became familiar with Gottstein's work. But the letters used here have proven to be good mnemonics for students, who have occasionally been confused by verticals coming before horizontals in the Gottstein system but after in standard sign lists.}, and compositions by various types of brackets.

\begin{table}[h]
\centering
\begin{tabular}{ccc}
\toprule
\textbf{Name} & \textbf{Example} & \textbf{Syntax} \\\midrule
Horizontal & \inlinesign{h} & \code{h} \\
Vertical & \inlinesign{v} & \code{v} \\
Downward Diagonal & \inlinesign{d} & \code{d} \\
Upward Diagonal & \inlinesign{u} & \code{u} \\
Hook & \inlinesign{c} & \code{c} \\\midrule
Horizontal Stack & \inlinesign{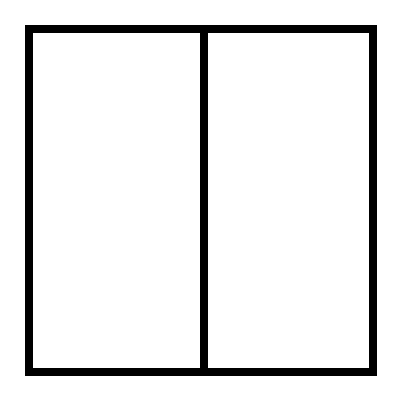} & \code{[]} \\
Vertical Stack & \inlinesign{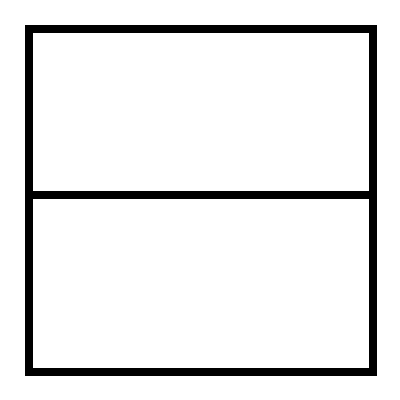} & \code{\{\}} \\
Intersect & \inlinesign{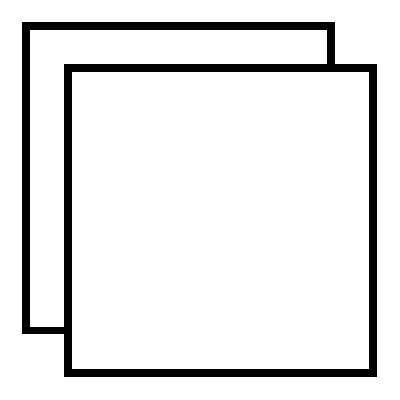} & \code{()} \\
\bottomrule
\end{tabular}
\vspace{0.5em}
\caption{The core of the \stelzer{} system.}
\label{tab:syntax}
\end{table}

These operations prove sufficient to describe almost all cuneiform signs used in Hittite. Hittite cuneiform is the particular focus of the initial work here for a few reasons: it has a relatively small sign inventory compared to most periods of Akkadian and Sumerian, and these signs were (in the words of \cite[29]{gordin2}) ``written with a keen sense of accuracy and symmetry'', with very reliable spacing and paragraph breaks\footnote{For the same reason, \citet{kriege} focus on it for their sign recognition experiments.}. At the same time, as discussed in section~\ref{subsec:lists}, the Hittite signs defy the straightforward classification of Neo-Assyrian, heightening the need for new indexing tools. And, last but certainly not least, ongoing Hittite classes at the University of Illinois offered a pool of students for testing. Extensions to other styles of cuneiform are discussed in section~\ref{subsec:aesthetic}.

\begin{figure}[h]
\begingroup
\Large
\begin{minipage}{0.30\linewidth}
\centering
\inlinesign{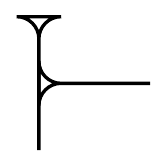} (\sign{me}) = \code{[vh]}

\begin{forest}
[\inlinesign{hstack}
	[\inlinesign{v}]
	[\inlinesign{h}]
]
\end{forest}
\end{minipage}
\begin{minipage}{0.30\linewidth}
\centering
\inlinesign{pa} (\sign{pa}) = \code{(v\{hh\})}

\begin{forest}
[\inlinesign{superpose}
	[\inlinesign{v}]
	[\inlinesign{vstack}
		[\inlinesign{h}]
		[\inlinesign{h}]
	]
]
\end{forest}
\end{minipage}
\begin{minipage}{0.38\linewidth}
\centering
\inlinesign{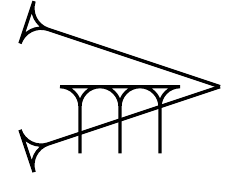} (\sign{ir}) = \code{\{d([vvv]u)\}}

\begin{forest}
[\inlinesign{vstack}
	[\inlinesign{d}]
	[\inlinesign{superpose}
		[\inlinesign{hstack}
			[\inlinesign{v}]
			[\inlinesign{v}]
			[\inlinesign{v}]
		]
		[\inlinesign{u}]
	]
]
\end{forest}
\end{minipage}

\endgroup
\caption{Examples of recursive encoding in the \stelzer{} system.}
\label{fig:coding}
\end{figure}

\subsection{Aesthetics}
\label{subsec:aesthetic}

\begin{figure}[h]
\centering
\includegraphics[width=0.5\linewidth]{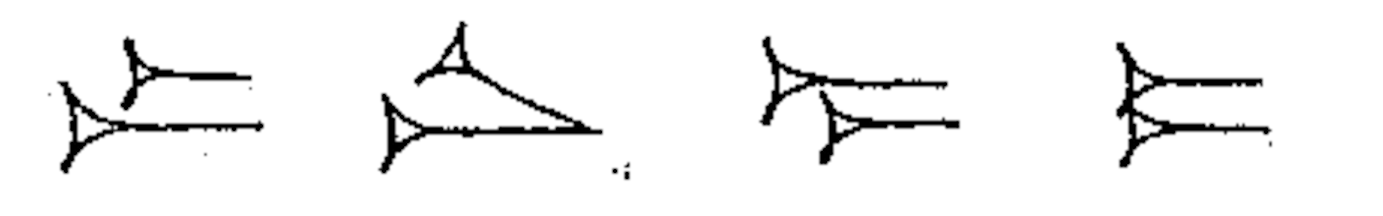}
\caption{Four consecutive headings from the \emph{Zeichenlexikon}'s sign index}
\label{fig:headings}
\end{figure}

The basic \stelzer{} system as presented in table~\ref{tab:syntax} only encodes the five basic types of strokes and their relationships to each other---not any other details of their position or size. This is a notable departure from previous systems like the indexing in \citet{hzl}, which considers a long horizontal stroke and a short horizontal stroke to be as thoroughly different as a horizontal and a diagonal\footnote{PaleoCodage similarly considers them fundamentally different, but for a more justifiable reason: the difference between them can be vitally important for palaeography.}. The four patterns shown in figure~\ref{fig:headings} are all separate headings in their sign index.

However, the main effect of this is to index many signs several times over. The sign \htt{pa} \inlinesign{pa}, for example, is listed separately under the first, third, and fourth of those headings---simply because scribes seldom seemed to notice or care which stroke was longer!

The \stelzer{} system is thus designed to ignore the exact size and placement of strokes in this encoding, focusing only on their relationships to each other. While the parallel strokes in \inlinesign{pa} may vary in length, they are always horizontal, and one is always placed above the other. This is intended to aid students in looking up signs, since they can ignore these details and focus only on the general pattern. During testing, multiple students expressed frustration with this particular aspect of traditional sign lists, saying they were unsure which of the headings from figure~\ref{fig:headings} they should be looking at.

\begin{wrapfigure}[11]{L}{0.4\textwidth}
\includegraphics[width=0.38\textwidth]{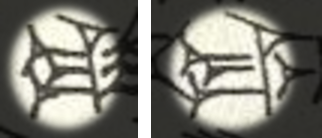}
\caption{Signs \htt{ku} (left) and \htt{ma} (right), from KBo 23.52}
\label{fig:kuma}
\end{wrapfigure}

It should be noted, however, that the lengths of strokes \emph{can} sometimes carry semantic meaning. The phonograms \htt{ku} and \htt{ma}, for example, are distinguished solely by the lengths of the horizontal strokes, as seen in figure~\ref{fig:kuma}. In the basic system, both would be encoded as \code{[\{hhh\}v]}. However, cases like this are rare, and even in these instances the stroke lengths can vary significantly. While this is a demonstrable limitation of the system, the practical impact is minimal: a student searching for one of these signs will find both, and decide based on context which is more appropriate.

These five basic strokes and three basic compositions are sufficient for a search system, but finer control of the details is important for other applications. A typesetting system, for example, \emph{must} be able to distinguish between \htt{ku} and \htt{ma}, and also make its middle stroke shorter than the outer ones (as in figure~\ref{fig:kuma}). And even for searching, students will---quite rightly---expect the search results to resemble what they see in the clay, rather than a Platonic version that ignores spacing and stroke length.

\begin{table}[h]
\centering
\begin{tabularx}{\textwidth}{cXc}
\toprule
\textbf{Name} & \textbf{Explanation} & \textbf{Syntax} \\\midrule
Void & A ``stroke'' that takes up space, either horizontally, vertically, or both, but is not displayed & \code{0} \\
Wildcard & A ``stroke'' indicating that \emph{something} stands in a particular position without committing to what it might be (for example, if a sign is damaged) & \code{*} \\
Shortening & Make a stroke shorter, either from the head end or the tail end & \code{'} (head), \code{"} (tail) \\
Doubling & Put a second head on the same stroke & \code{2} \\
Damage & Hatch over a part of the sign to indicate damage to the tablet & \code{\#} \\
Expansion & Instruct the renderer to give this component more space & \code{E} \\
Restriction & Prevent a component from expanding in any direction & \code{R} \\
Margin & Instruct the renderer to leave some empty space around this component & \code{M} \\
\emph{Tenû} (or Tilt) & Rotate an entire component 45 degrees & \code{T} \\
Canvas size & Change the width of the canvas & (Various) \\
\bottomrule
\end{tabularx}
\vspace{0.5em}
\caption{Advanced syntax}
\label{tab:advsyntax}
\end{table}

For these reasons, the system includes a set of \emph{modifiers}, shown in table~\ref{tab:advsyntax}. These can be used to lay out a sign in finer detail\footnote{Though significantly less fine of detail than PaleoCodage.}, adjusting the size and positioning of strokes within the recursive structure; the effects can be seen in figure~\ref{fig:complex}. These are removed during normalization (see section~\ref{subsec:normalize}), ensuring that they're ignored while searching---they are useful for rendering, but students don't need to use them to find the sign they need.

\begin{figure}[h]
\includegraphics[width=\linewidth]{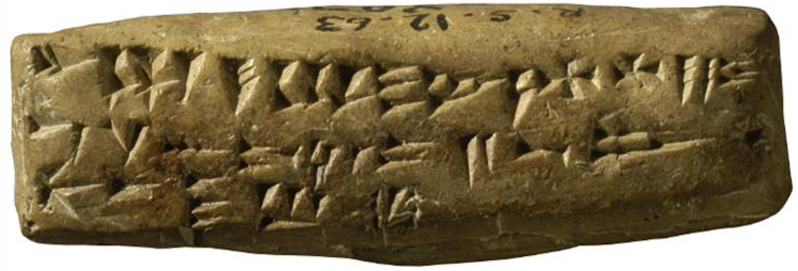} \\
\includegraphics[width=\linewidth]{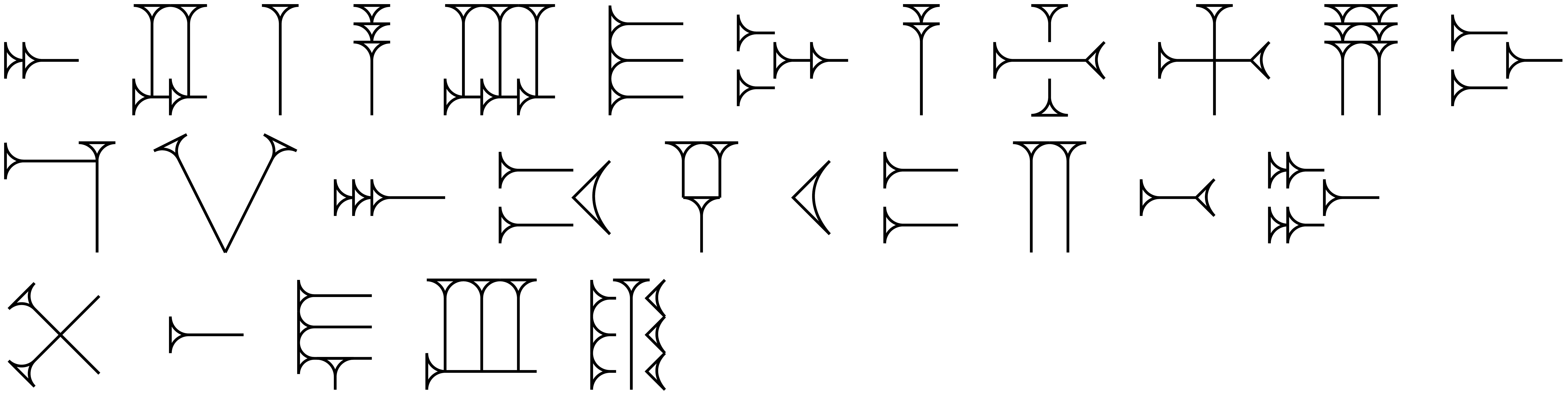}

\caption{A Ugaritic abecedary tablet (RS 12.063/KTU 5.6) rendered using the prototype system, as a demonstration of non-Hittite cuneiform. The \code{3} (triple-headed) and \code{?} (inverted) modifiers are required here. Tablet photo originally from \citet[124, figure 2a]{yon}, reproduced at \url{https://mnamon.sns.it/index.php?page=Esempi&id=30&lang=en}.}
\label{fig:ugaritic}
\end{figure}

Using these modifiers, the \stelzer{} system can easily be extended to other eras and styles of cuneiform. For example, Old Babylonian cursive (unlike Hittite) sometimes makes use of upward vertical strokes, as in the sign \emph{nu}. But these are far less common than the five types of strokes given above---rare enough that they don't receive their own headings when categorizing signs, or their own codes in Gottstein or PaleoCodage. To add these strokes to the system, all that is needed is a new modifier, \code{?}, which inverts the direction of a stroke\footnote{Akin to the \texttt{!} modifier in PaleoCodage.}. The Old Babylonian cursive \emph{nu} can then be encoded as \code{(hv?)}, producing \inlinesign{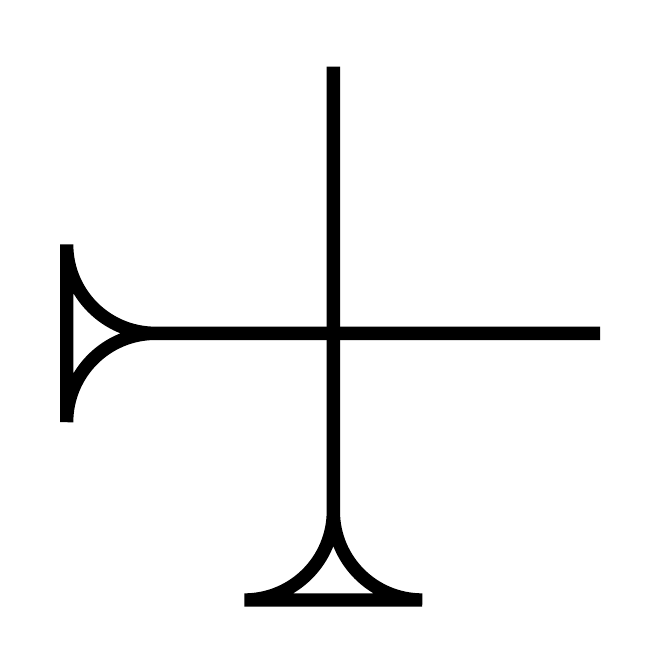}. Similarly, the triple-headed strokes occasionally used in Assyrian can be implemented with a new \code{3} modifier; Old Assyrian \smr{bal} `libation' \inlinesign{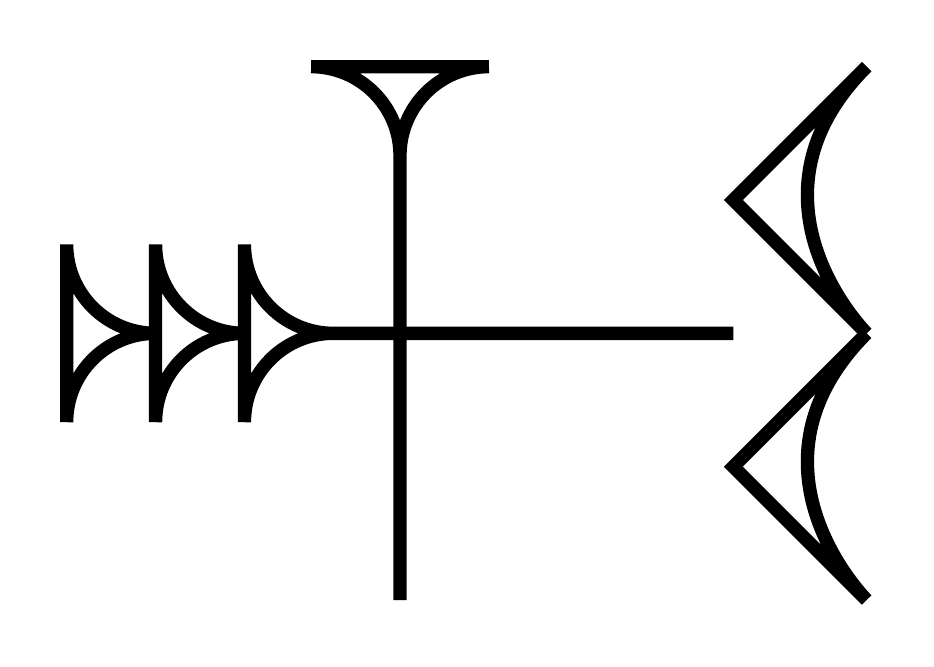} becomes \code{[(h3v)\{cc\}]}. \posscite{paleocodage} seal wedges could be implemented in the same way. For a fuller demonstration, see figure~\ref{fig:ugaritic}.

\begin{table}[p]
\begin{tabularx}{\linewidth}{cccX}
\toprule
\multirowcell{9}{\textbf{Strokes}} & \code{h} & Horizontal & \inlinesign{h} \\
& \code{v} & Vertical & \inlinesign{v} \\
& \code{d} & Downward Diagonal & \inlinesign{d} \\
& \code{u} & Upward Diagonal & \inlinesign{u} \\
& \code{c} & Winkelhaken/Hook & \inlinesign{c} \\
& \code{0} & Void & An invisible ``stroke'' that takes up space \\
& \code{*} & Wildcard & Matches any stroke in the encompassing algorithm \\
& \code{|} & Cursor & Renders as a line or cross, to show where a new stroke will be inserted in the interface; ignored in searching \\ \midrule
\multirowcell{3}{\textbf{Compositions}} & \code{[]} & Horizontal Stack & Arrange children from left to right \\
& \code{\{\}} & Vertical Stack & Arrange children from top to bottom \\
& \code{()} & Superposition & Overlay children onto the same space \\ \midrule
\multirowcell{11}{\textbf{Stroke}\\\textbf{Modifiers}} & \code{'} & Shorten Head & For most strokes, shorten by bringing the head inward; for a Winkelhaken, reduce size slightly; for a void, prevent expanding horizontally \\
& \code{"} & Shorten Tail & For most strokes, shorten by bringing the tail inward; for a Winkelhaken, reduce size greatly; for a void, prevent expanding vertically \\
& \code{2} & Double Head & Put an additional head on the stroke \\
& \code{3} & Triple Head & Put two additional heads on the stroke (not used in Hittite, needed for Old Assyrian) \\
& \code{\#} & Damage & Draw hatching (diagonal lines) over this stroke, to indicate damage to the tablet in rendering \\
& \code{!} & Highlight & Render this stroke in a different color \\
& \code{?} & Invert & Swap the head and tail of this stroke (not used in Hittite, needed for Old Babylonian) \\
\midrule
\multirowcell{6}{\textbf{Node}\\\textbf{Adjustments}} & \code{T} & \emph{Tenû} & Rotate a node 45 degrees counter-clockwise \\
& \code{E} & Expand & Ask the arrangement algorithm to give this node twice as much space \\
& \code{M} & Margin & Leave a small amount of empty space on all sides of this node \\
& \code{R} & Restrict & Prevent this node from expanding in any direction \\ \midrule
\multirowcell{6}{\textbf{Canvas}\\\textbf{Size}} & \code{N} & Narrow & 1:3 (width:height ratio) \\
& \code{P} & Portrait & 2:3 \\
& \code{S} & Square & 1:1 (default) \\
& \code{L} & Landscape & 3:2 \\
& \code{W} & Wide & 2:1 \\
& \code{X} & Extra-wide & 3:1 \\
\bottomrule
\end{tabularx}
\vspace{0.5em}
\caption{A complete syntax reference. Modifiers and adjustments come after the nodes they modify; commas and whitespace can be used as optional delimiters.}
\label{tab:ref}
\end{table}

\subsection{Semantics}

Some of the modifiers in table~\ref{tab:advsyntax} require a bit more attention. Many of them are purely aesthetic, and can be simply removed without changing the fundamental nature of the sign. Others require a bit of adjustment when they're removed: \code{v2} (a double-headed vertical stroke) is more similar to \code{\{vv\}} than \code{v}, two verticals rather than one, but the difference between \code{v2} and \code{\{vv\}} is purely an aesthetic one.

The \emph{tenû} modifier, though, is a fundamental change to the sign: \inlinesign{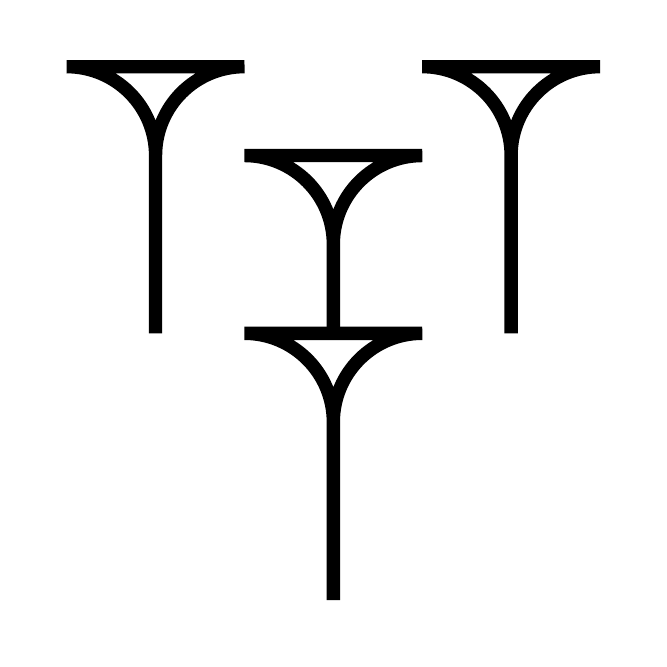} \smr{ninda} `bread' is a different sign from \inlinesign{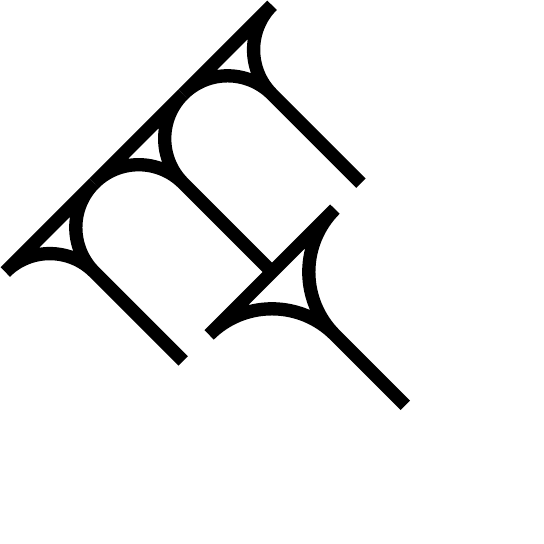} \htt{\heth{}i}. Why should this be implemented as a modifier? Wouldn't it be better to have a ``diagonal stacking'' composition to capture cases like \inlinesign{hi}, encoded perhaps as \code{\textlangle{}<ddd>d\textrangle{}}---akin to PaleoCodage's \texttt{.} and \texttt{,} operators?

Notably, though, this ``diagonal stacking'' has a very restricted distribution. While diagonal strokes are frequently stacked horizontally and vertically (e.g. \htt{ni} \inlinesign{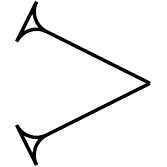}), horizontal and vertical strokes are never stacked diagonally. And with only one possible exception\footnote{See section~\ref{subsec:exceptions}.}, horizontal and vertical stacks never appear inside or overlap with diagonal ones in Hittite.

Instead, components (or even entire signs) are written ``\emph{tenû}'': rotated by 45 degrees\footnote{See \citet{ice2} for examples.}. It is likely that this was handled by physically rotating the tablet, as opposed to horizontal and vertical strokes made using different edges of the stylus. In other words, these ``diagonal stacks'' are fundamentally a modification of horizontal and vertical stacks, rather than a separate entity of their own.

For these reasons, diagonal stacking is not considered its own separate composition in the system. Rather, it is treated as either a horizontal or vertical stack, containing only horizontal and vertical strokes, with the \emph{tenû} modifier applied. During normalization, horizontal strokes within a \emph{tenû} composition are replaced with upward diagonals, while vertical strokes become downward diagonals\footnote{When talking about entire signs, \emph{tenû} generally means a clockwise rotation. Here it is instead implemented counterclockwise, to avoid needing to reverse the directions of strokes: vertical strokes \emph{tenû} generally point to the right rather than to the left.}---meaning that a search for \inlinesign{hi} won't find \inlinesign{ninda}.

\subsection{Exceptions}
\label{subsec:exceptions}

The first true test of this system is its completeness. With the modifiers from table~\ref{tab:advsyntax}, it can make a satisfactory rendering of all but two signs in the \emph{Hethitisches Zeichenlexikon}. These two exceptions are the logograms \smr{ga\shin{}an} `Lady' (a title used in prayers to female deities) and \smr{dal\heth{}amun\textsubscript{4}} `Dalhamun' (an epithet of the Akkadian storm deity Adad).

\begin{wrapfigure}[27]{R}{0.4\textwidth}
\includegraphics[width=0.38\textwidth]{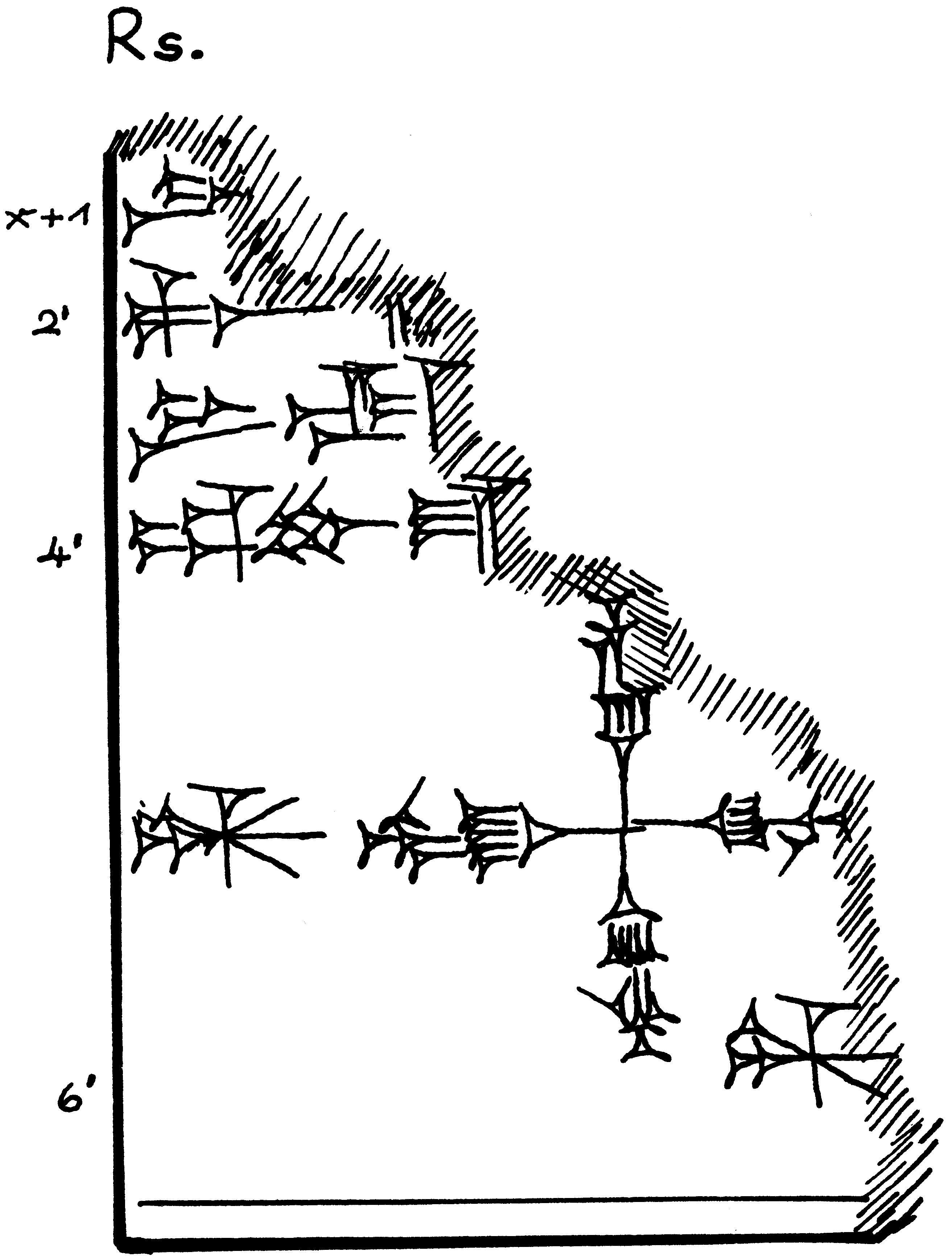}
\caption{A Kreuzform sign, ``\sign{\shin{}ir}\x4''. Autograph originally from KBo \citep{kbo}; image taken from \citet{torri}.}
\label{fig:kreuzform}
\end{wrapfigure}

The latter can be quite reasonably ignored. It consists of the sign \sign{naga} written four times in different directions, converging on a center point to make a cross shape. A similar construction (shown in figure~\ref{fig:kreuzform}) involves four copies of the logogram \smr{\shin{}ir} `testicle' arranged in the same pattern, likely an epithet of the moon-god; this one is a hapax and doesn't even merit an official name or Unicode codepoint.

These ``Kreuzform'' signs don't fit into a standard line of text, and in fact never seem to be used inline on actual tablets in the Hittite era: they appear only in lists of logograms or sketched in colophons \citep{torri}. They are not given their own index numbers in \citet{hzl} and are not supported in any common Hittite fonts, so I have no qualms about leaving them unencoded\footnote{\smr{dal\heth{}amun\textsubscript{4}} is considered iconic enough to appear on the cover of the \emph{Zeichenlexikon}, but as a symbol, not as a meaning-bearing logogram.}.

\begin{figure}[h]
\centering
\includegraphics[width=0.67\linewidth]{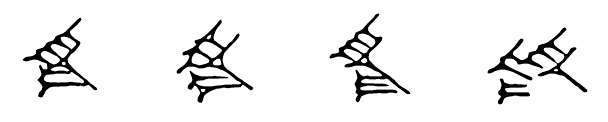} \\
\includegraphics[width=0.67\linewidth]{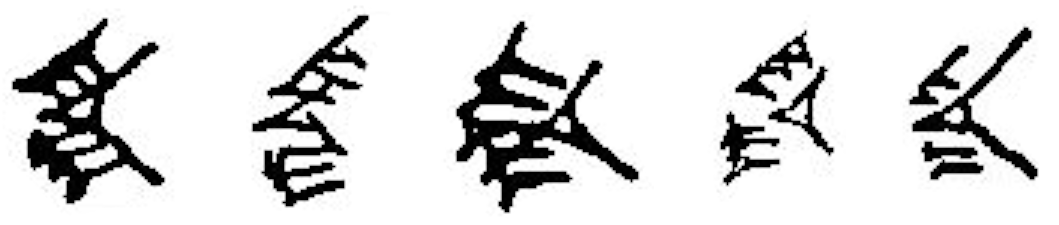} \\
\includegraphics[width=0.67\linewidth]{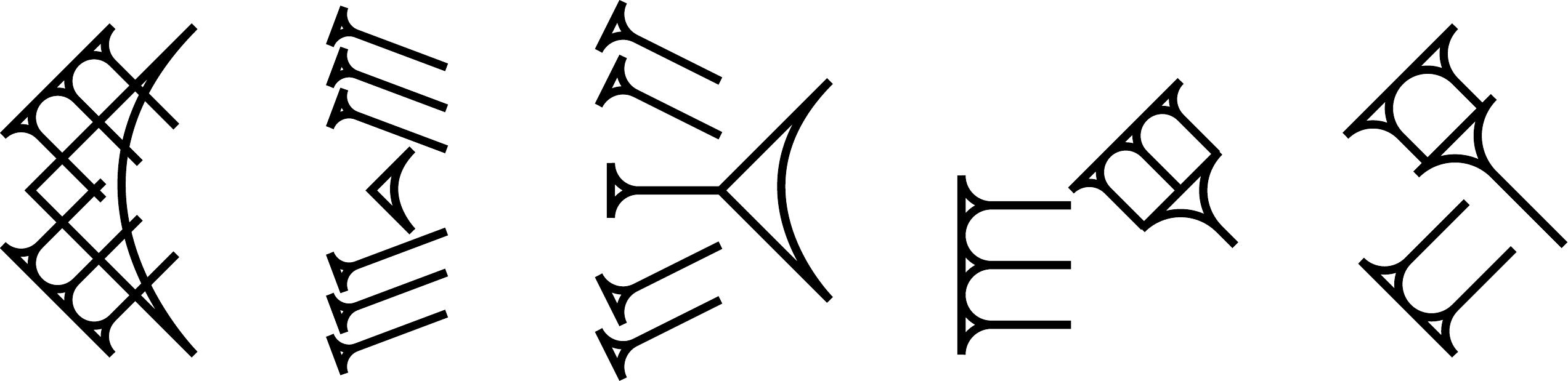}
\caption{Variants of the sign \smr{ga\shin{}an} `Lady'. At the top, how it's presented in the \emph{Zeichenlexikon}; in the middle, how it appears on actual tablets (KBo 8.110, KBo 24.43, KBo 47.133, KUB 27.1, and KBo 39.159); on the bottom, different ways of representing it in the \stelzer{} system.}
\label{fig:gacan}
\end{figure}

The sign \sign{ga\shin{}an} poses a larger issue. As presented in the \emph{Zeichenlexikon}, it involves horizontal strokes meeting a diagonal, as shown at the top of figure~\ref{fig:gacan}. This is something the \stelzer{} system currently can't handle---the best way to do it would be to put downward diagonals inside a \emph{tenû} modifier, which is not allowed.

Curiously, though, the sign as written on actual tablets tends to look quite different. Several examples can be seen in the middle of figure~\ref{fig:gacan}, and none of them quite match the \emph{Zeichenlexikon}'s form. This makes sense, if diagonal strokes were indeed made by rotating the tablet: having horizontal strokes truly \emph{meet} a diagonal would require significantly more effort than meeting a vertical.

Historically speaking, this sign started as a \emph{gunû} (``decorated'') form of the sign \smr{u} `ten', a single Winkelhaken. This can be represented in the \stelzer{} system by superposing \emph{tenû} strokes (bottom left). Various other representations are possible too, as shown across the bottom; while none of them are perfect representations of how it's written on the tablets, they can approximate it significantly better than they can the \emph{Zeichenlexikon} form. The question of what the Platonic ideal of the Hittite \sign{ga\shin{}an} should truly look like is left for future work.

\subsection{Comparison}

The benefits over the Gottstein system are clear. The \stelzer{} encoding includes the same information about the stroke types, plus additional information about how they relate; a database of signs in the \stelzer{} system can be trivially converted to the Gottstein system, or allow searching by Gottstein code as an alternative.

\begin{table}[h]
\centering
\begin{tabular}{ccccc}
\toprule
\textbf{Sign} & \textbf{Gottstein} & \textbf{PaleoCodage} & \textbf{\stelzer{}} \\\midrule
{\Large \inlinesign{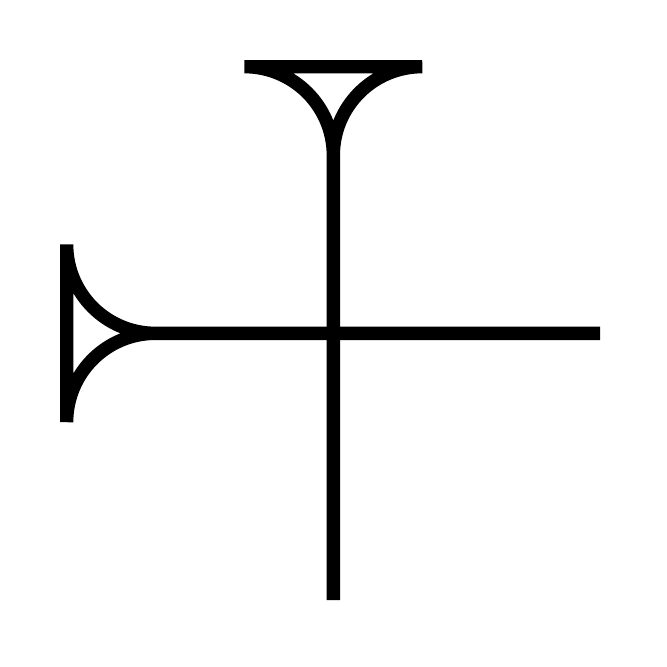}} \sign{ma\shin} & \texttt{a1b1} & \texttt{:b-a} & \code{(vh)} \\
{\Large \inlinesign{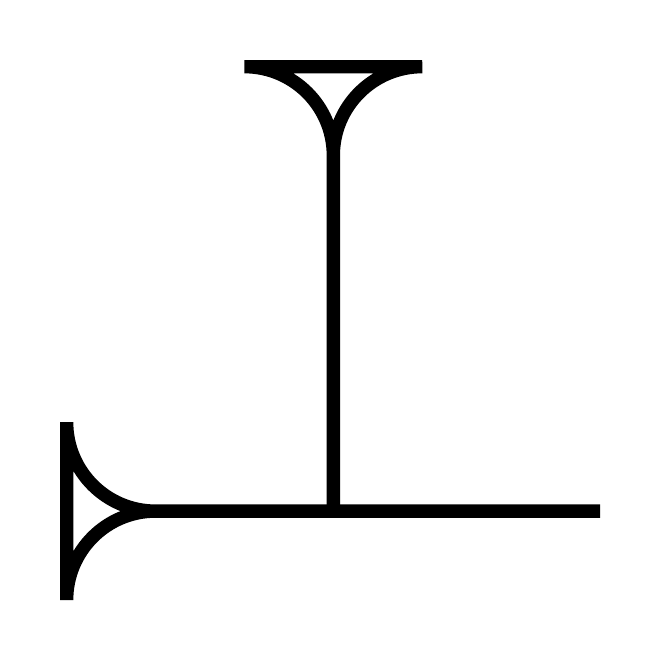}} \sign{bar} & \texttt{a1b1} & \texttt{;b-a} & \code{\{vh\}} \\
{\Large \inlinesign{me}} \sign{me} & \texttt{a1b1} & \texttt{a-:b} & \code{[vh]} \\
\bottomrule
\end{tabular}
\vspace{0.5em}
\caption{Three signs\protect\footnotemark{} with identical Gottstein codes, but distinct PaleoCodage and \stelzer{} encoding.}
\label{tab:compmini}
\end{table}

\footnotetext{This demonstration is taken from \citet[ii131]{paleocodage}, with the \stelzer{} encoding chosen to match the PaleoCodage forms; in Hittite, the signs \sign{ma\shin} and \sign{bar} have merged.}

The benefits over PaleoCodage are less obvious. Both of them fundamentally try to represent the same thing: the relative positions of strokes within a sign. What is there to gain by representing this with a tree, rather than state changes in a finite automaton?

The key is that, in the \stelzer{} encoding, the relationships between the strokes themselves are fundamental. In PaleoCodage, the operator between two strokes doesn't just express the relationship between them---it conveys how they relate to the sign as a whole, and the other strokes in it. As mentioned in section~\ref{subsec:otherenc}, the difference between \texttt{a-a} and \texttt{a\_a} (a vertical stroke next to a vertical stroke) doesn't necessarily represent anything about those two strokes. Instead, the primary difference is how they interact with other types of strokes overlapping them.

In the \stelzer{} system, on the other hand, the relationship between \emph{any} two strokes can be determined by finding their last common ancestor in the tree. In the sign \inlinesign{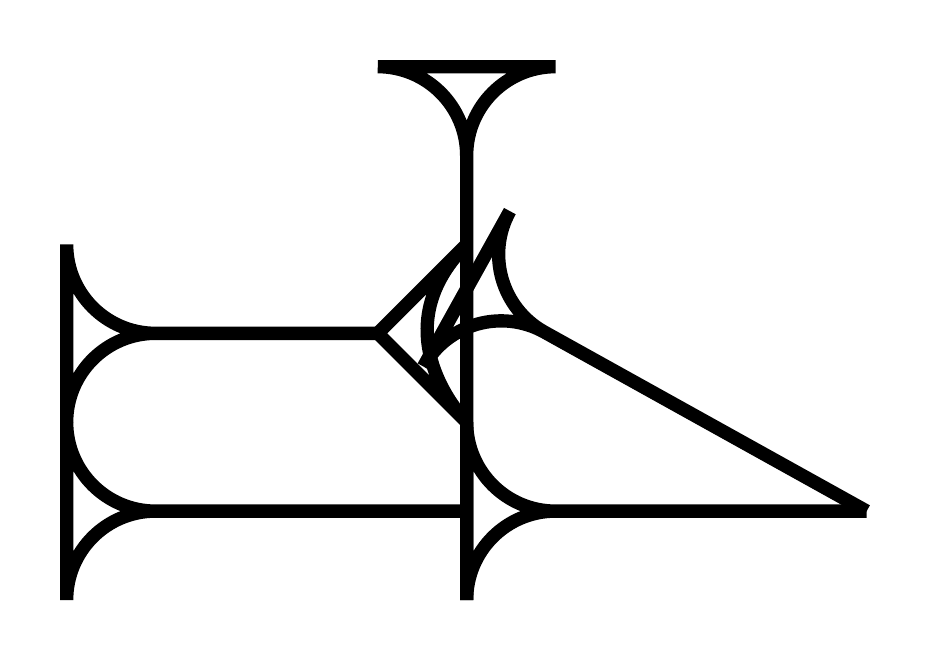} \smr{ge\shin{}tin} `wine', the hook stroke is to the left of the diagonal stroke---and this is reflected in the \stelzer{} encoding (as shown in figure~\ref{fig:gectin}). The last common ancestor of those two (shown by the blue lines) is a horizontal stack, and the hook precedes the diagonal.

The encompassing algorithm described in section~\ref{subsec:encompass} then ensures that a search for \code{[cd]} will find this sign, no matter how far apart the hook and diagonal are in the encoded sign. As a result, a student using the \stelzer{} encoding can search for \emph{any part of a sign}---no matter whether other parts of the sign are missing, damaged, badly transcribed, or just difficult to encode. This is something that neither the Gottstein nor PaleoCodage systems currently supports. The \stelzer{} system is designed around the relationships between strokes first and foremost, and as a result is far more resilient to damage or just simple obscurity than its predecessors.

\begin{figure}[h]
\begingroup
\Large
\begin{minipage}{\linewidth}
\centering
{\Huge\inlinesign{gectin}} \code{[\{0[hc]h\}v\{0dh\}]}

\begin{forest}
[\inlinesign{hstack}
	[\inlinesign{vstack}, edge={blue,very thick}
		[\code{0}]
		[\inlinesign{hstack}, edge={blue,very thick}
			[\inlinesign{h}]
			[\inlinesign{c}, edge={blue,very thick}]
		]
		[\inlinesign{h}]
	]
	[\inlinesign{v}]
	[\inlinesign{vstack}, edge={blue,very thick}
		[\code{0}]
		[\inlinesign{d}, edge={blue,very thick}]
		[\inlinesign{h}]
	]
]
\end{forest}
\end{minipage}
\endgroup
\caption{The relationships between certain strokes in the sign \smr{ge\shin{}tin} `wine', highlighted in blue.}
\label{fig:gectin}
\end{figure}

A comparison of how a few different signs would be encoded in these three systems can be found in table~\ref{tab:compmini}, and a fuller demonstration in table~\ref{tab:compare}.

\begin{table}[h]
\centering
\begin{tabular}{cccccc}
\toprule
Sign & Name & Gottstein & \stelzer{} & PaleoCodage & PaleoCodage Render \\
\midrule
{\Huge\inlinesign{pcx_ya}} & \htt{ya} & \texttt{a3b5} & \code{L[\{hhh\}\{hh\}vv2]} & \texttt{b:b:b\_/b:b\_a-a:a} & {\LARGE\inlinesign{pc_ya}} \\
{\Huge\inlinesign{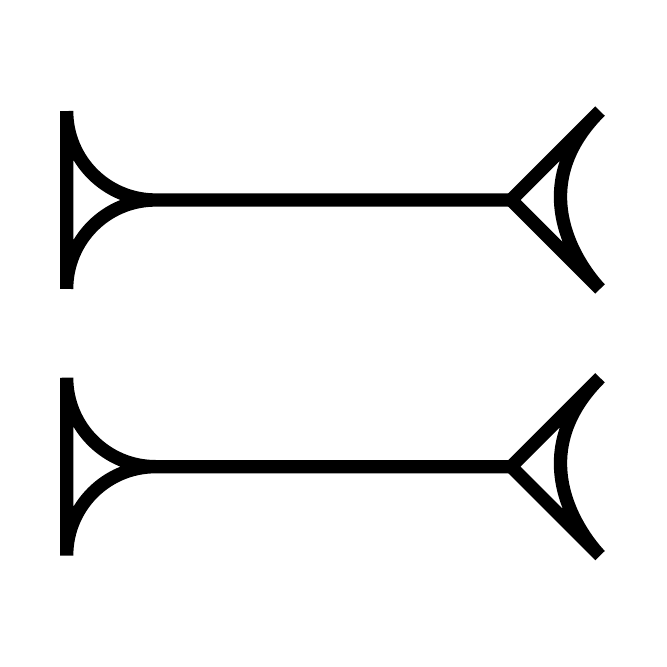}} & \htt{bi} & \texttt{b2c2} & \code{\{[hc'][hc']\}} & \texttt{b:b\_w:w} & {\large\inlinesign{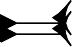}} \\
{\Huge\inlinesign{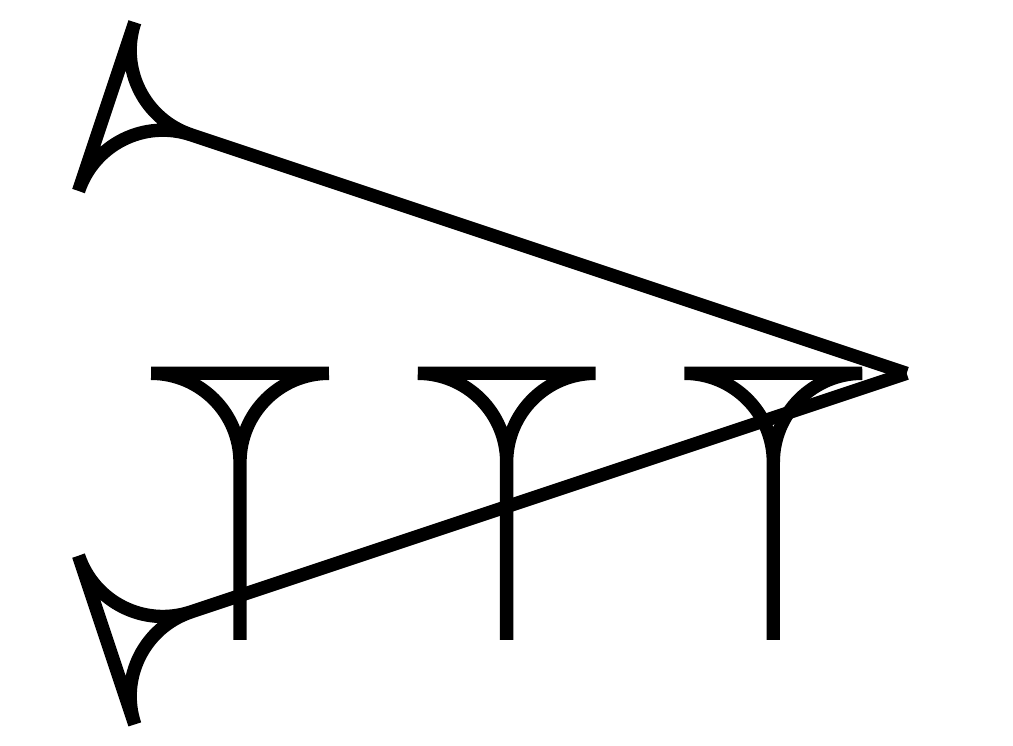}} & \htt{ir} & \texttt{a3c2} & \code{L\{d(u[vvv])\}} & \texttt{<;C>{}>D-:;sa-:;sa-:;sa} & {\Huge\inlinesign{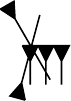}} \\ %
{\Huge\inlinesign{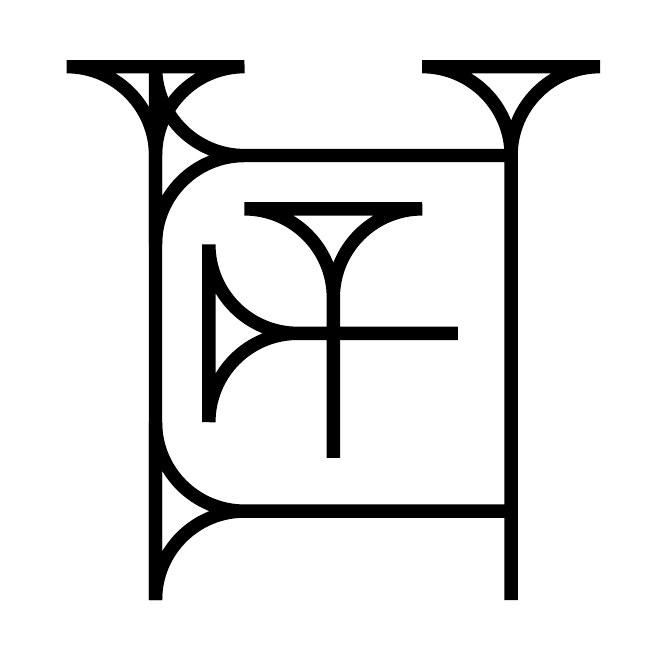}} & \htt{lu} & \texttt{a3b3} & \code{[v\{h(hv)Mh\}v]} & \texttt{:a-b;b-:::sb::-::sa-a} & {\Huge\inlinesign{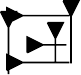}} \\
\bottomrule
\end{tabular}
\vspace{0.5em}
\caption{Various phonetic signs encoded in each system, for comparison. Gottstein code, PaleoCodage encoding, and PaleoCodage rendering (the rightmost column) are taken from the demonstration at \url{https://situx.github.io/PaleoCodage/}. \stelzer{} encoding and rendering (the leftmost column) are adapted to match this version of the sign if it differs from the Hittite one.}
\label{tab:compare}
\end{table}

\section{Algorithms}
\label{sec:algo}

\subsection{Rendering}
\label{subsec:render}

Now that we have a way of encoding signs into trees, we can write algorithms to manipulate them in various ways. For example, we can render a tree back into an image of a sign; see \code{elements.py} in the provided code for the implementation. Unlike in PaleoCodage, the encoding is not expected to specify the exact size, position, and angle of each sign---instead, the rendering algorithm applies various typographical and aesthetic principles to arrange the strokes in a clear and pleasing way. This means that, for example, diagonal strokes in the \stelzer{} system only encode whether they're oriented upward or downward; the exact angle is chosen by the rendering algorithm based on aesthetic constraints.

The first step is assigning each node in the tree a certain amount of space (represented as a rectangle\footnote{Or, in computer graphics terms, an ``axis-aligned bounding box'' encoding width, height, and the coordinates of the top-left corner, but not rotation.}). The root node is given the entire canvas; branching nodes divide up their space among their children, while non-branching nodes take whatever they're given. Horizontal stacks divide up the rectangle horizontally; vertical stacks divide up the rectangle vertically; superpositions give the entire space to each child, causing them to intersect. Strokes then render themselves into their rectangles.

The main difficulty is apportioning the space in horizontal and vertical stacks. As a first approximation, the stack divides up the space evenly between its children, giving twice as much space to any child that's an \code{E} adjustment. It then checks whether any child is able to expand in the direction of stacking. For example, \inlinesign{h} strokes can expand horizontally, but cannot expand vertically, while \inlinesign{v} strokes can expand vertically but not horizontally, \inlinesign{d} can expand in both directions, and \inlinesign{c} cannot expand at all. A composition can expand if any of its children can expand.

If any of the stack's children can expand in the appropriate direction, the program checks how much space the non-expanding children are actually using, recording how much of their allotted space is currently going unused, and how much space can be reclaimed through kerning: each node is given the task of determining how far it can be kerned into in each direction. Then this additional space is evenly distributed between the expandable children, and all the children are repositioned appropriately.

\begin{wrapfigure}[15]{L}{0.4\textwidth}
\includegraphics[width=0.19\textwidth]{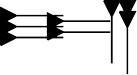} \hfill \includegraphics[width=0.19\textwidth]{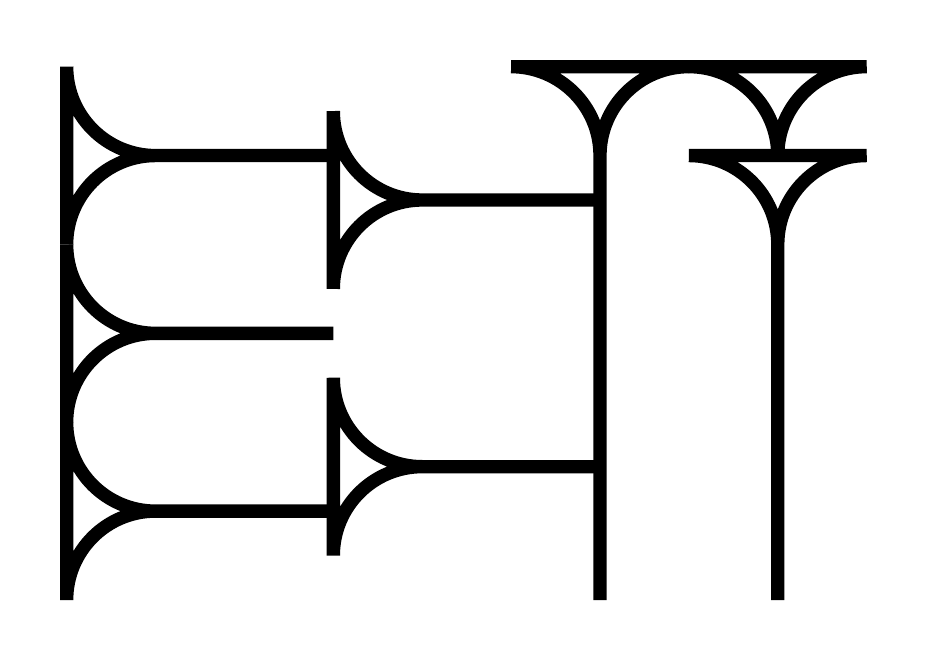}
\caption{The phonetic sign \htt{ya}, rendered in PaleoCodage (left) and the \stelzer{} system (right). PaleoCodage's renderer is optimized for precision in detail, as necessary for palaeography, while \stelzer{}'s is optimized for overall aesthetics and readability at the expense of precision.}
\label{fig:pcrender}
\end{wrapfigure}

\begin{figure}[h]
\centering
\includegraphics[width=\linewidth]{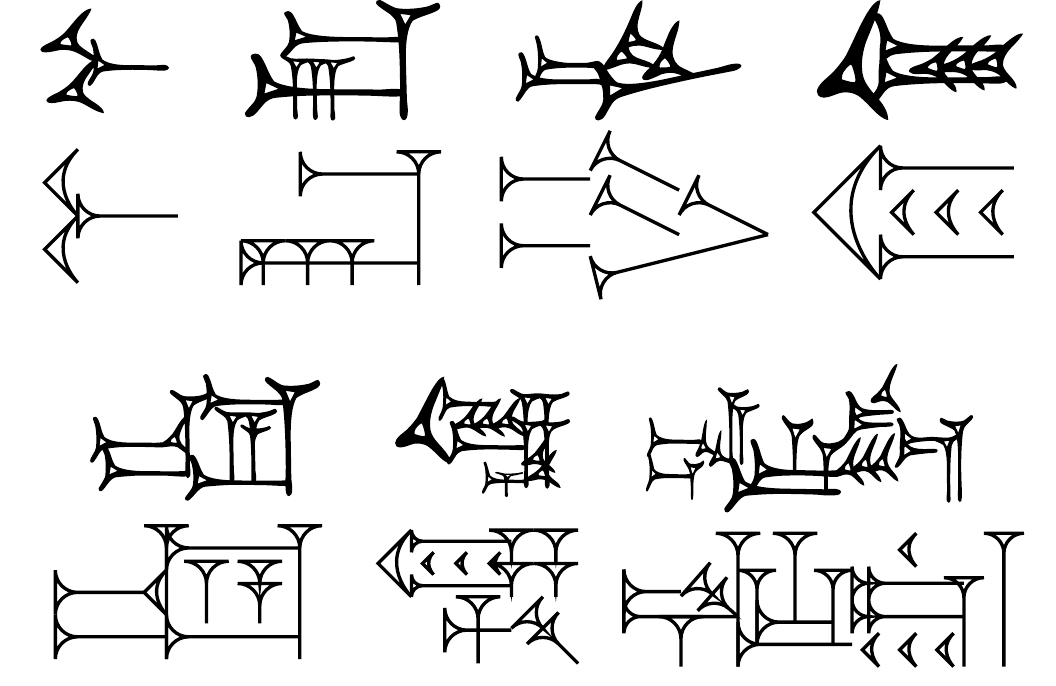}
\caption{Logograms of varying complexity rendered in a traditional hand-drawn font (lines 1 and 3) and in our new rendering engine (lines 2 and 4). Top row: \smr{munus} `woman', \smr{iku} `field', \smr{lugal} `king', \smr{ki\shin{}} `world'. Bottom row: \smr{nag} `drink', \smr{buru\textsubscript{14}} `harvest', \smr{umbin} `fingernail'. Font used: \emph{Ullikummi}, created by Sylvie Vanséveren, based on \citet{hzl}.}
\label{fig:font}
\end{figure}

This algorithm is simple---requiring much less detail than the palaeologically-focused PaleoCodage---but the end results are remarkably effective. With only two exceptions (see section~\ref{subsec:exceptions}), it has been able to create aesthetically pleasing renditions of every sign and common variant listed in \citet{hzl}. Figure~\ref{fig:pcrender} compares the results against the detail-oriented PaleoCodage renderer, and figure~\ref{fig:font} compares this algorithm's rendering of several logograms against a traditional hand-drawn cuneiform font.

While the original goal was only to encode signs for searching purposes, this renderer compares favorably against traditional fonts. It could be used to quickly render unusual glyphs and glyph variants for discussion, much like \posscite{wong} proposed HanGlyph synthesizer\footnote{See figure 6 in \citet[587]{wong} for an example.}, and could potentially render entire documents in a clean, easily-readable style. All the inline glyphs used in this article are rendered in this way. While PaleoCodage can be far more precise in the details---even with aesthetic modifiers, the \stelzer{} system is not designed to capture the palaeographic details of a particular scribe's handwriting---the output of the \stelzer{} renderer is meant to be more readable for general usage, equivalent to a typeset edition of a manuscript rather than a facsimile.

\begin{figure}[h]
\includegraphics[width=\linewidth]{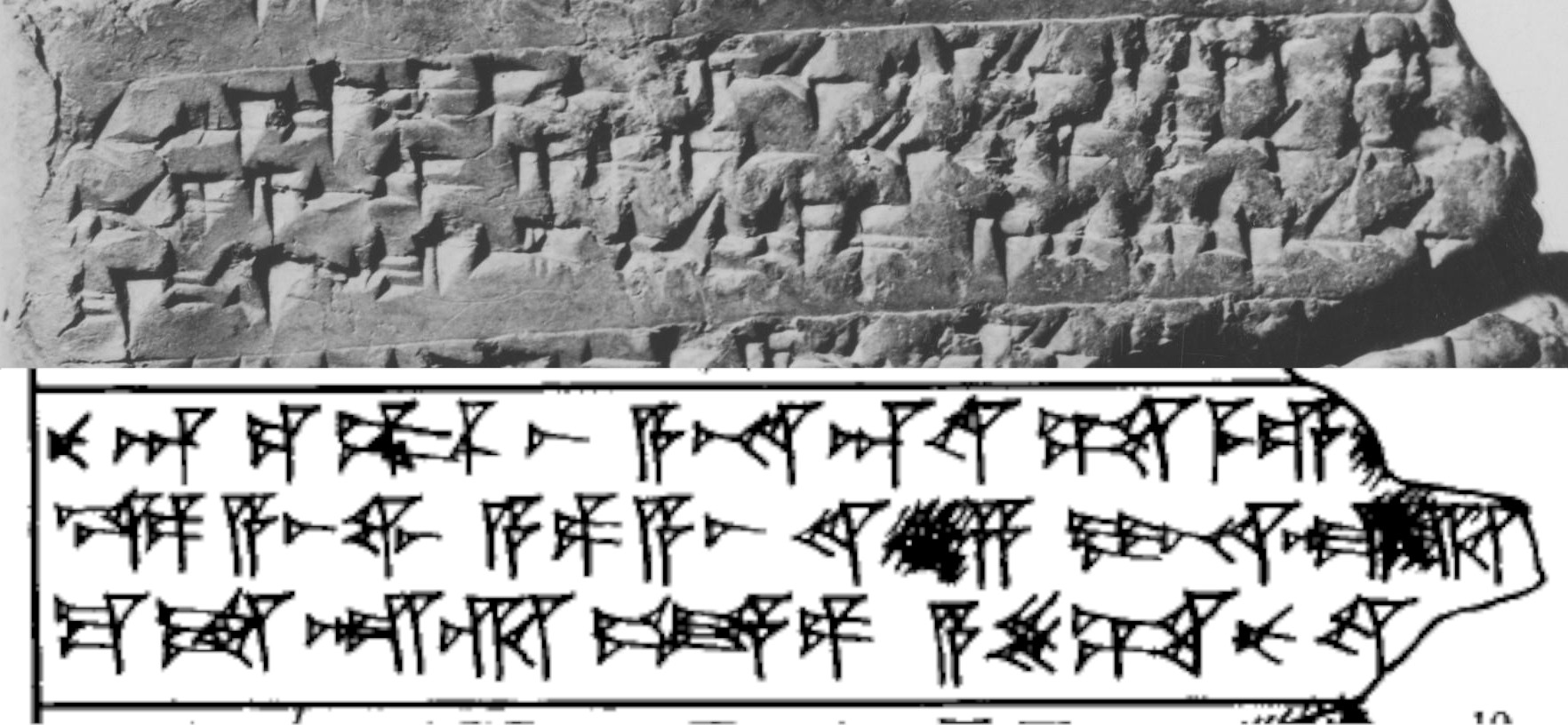} \\
\includegraphics[width=\linewidth]{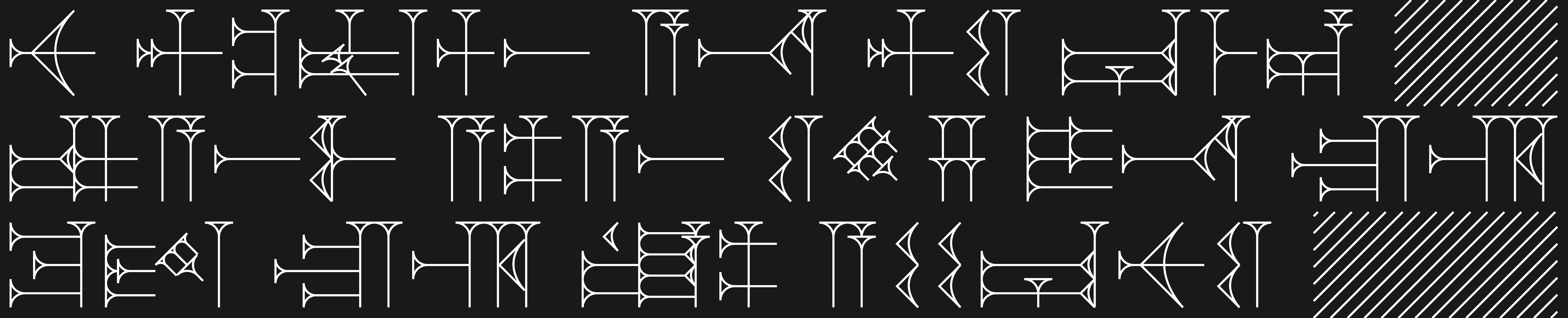}

\caption{A prototype rendering of three lines from a tablet---in this case, the Hittite Gilgamesh. Unlike PaleoCodage, the \stelzer{} system renderer is optimized for overall readability, sacrificing palaeographic detail for this purpose (e.g. note the altered shape of the \sign{an} sign, second in the first line). Autograph taken from \url{https://www.assyrianlanguages.org/hittite/index_en.php?page=textes}.}
\label{fig:tabletrender}
\end{figure}

Hittite cuneiform tablets in particular have a reputation for being very clean and readable in their layout, even in daily correspondence: words are clearly distinguished, lines of text are more or less justified, paragraphs and sections are marked. Some cuneiform scholars\footnote{\citeauthor{gordin2} cites Joachim Marzahn.} will even ``lovingly'' call it ``typewriter cuneiform'' (\emph{Schreibmaschinen-Keilschrift}) \citep[27--29]{gordin2}. Could the renderer replicate these features, laying out full tablets on these principles? This would provide, effectively, cuneiform typesetting of a sort seldom attempted before. A prototype can be seen in figure~\ref{fig:tabletrender}.

\subsection{Normalization}
\label{subsec:normalize}

One issue with this encoding is that it can be ambiguous. While every recursive code describes exactly one sign, the same sign can sometimes be described by multiple codes. \inlinesign{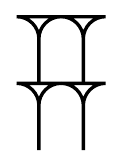} \htt{za}, for example, could be described either as \code{[\{vv\}\{vv\}]} or as \code{\{[vv][vv]\}}: a horizontal stack of two vertical stacks, or a vertical stack of two horizontal stacks. This poses a problem for searching. How can a student know which of these two they should search for?

The solution is a recursive algorithm for \emph{normalization}: converting encodings to a form that is easy and unambiguous to compare. The main purpose of this ``normalized'' or ``functional'' form is to be used as the input to other algorithms. As a proof of concept, two different ``modes'' of normalization are implemented; in the standard mode, all five stroke types remain distinct, but in ``Gottstein mode'', downward diagonals and Winkelhaken are not distinguished, as recommended by \citet{gottstein}. This demonstrates that similar adjustments could be made for particular languages or eras as necessary.

In the following algorithms, the word \emph{contains} means ``is the parent of'', while \emph{indirectly contains} means ``is the ancestor of''. That is, a node contains its children, and indirectly contains its children, its children's children, and so on.

The normalization algorithm is as follows:
\begin{itemize}
\item The normalized form of a double-headed stroke is a stack of two single strokes (\code{h2} \textrightarrow{} \code{[hh]})\footnote{Likewise for triple-headed strokes---not part of the system for Hittite, but discussed in section~\ref{subsec:aesthetic} as it pertains to Old Assyrian.}.
\item If operating in ``Gottstein mode'', the normalized form of a downward diagonal is a Winkelhaken, or a vertical stack of two Winkelhaken if double-headed. This normalizes away the difference between these two types of strokes for styles of cuneiform where they are not reliably distinguished.
\item The normalized form of a void is nothing at all. In other words, voids are discarded.
\item The normalized form of any other stroke is that stroke without any modifiers.
\item The normalized form of the \emph{tenû} adjustment is the normalized form of its child, with all horizontals replaced by upward diagonals, and all verticals replaced by downward diagonals\footnote{Internally, this is accomplished by setting a flag that's propagated through the recursive algorithm, using the same mechanism as the mode flag.}.
\item The normalized form of any other node adjustment (like \code{E}) is the normalized form of its child.
\item Normalizing a composition starts by calculating the normalized forms of all its children. If it's a superposition (where the order of the children doesn't matter), the children are sorted in lexicographic order\footnote{That is, \code{c} comes before \code{h} comes before \code{v}. This is entirely arbitrary, but consistent.}. \par Then, a few special cases are checked:
\begin{itemize}
	\item If a composition has only a single child, the normalized form of the composition is the normalized form of the child (\code{[v]} \textrightarrow{} \code{v}).
	\item If a composition has no children, the normalized form of that composition is nothing at all.
	\item If a composition contains another composition of the same type, the nesting is removed (\code{[v[vv]v]} \textrightarrow{} \code{[vvvv]}).
	\item If a vertical stack contains only horizontal stacks, and all those horizontal stacks have the same number of elements, and the parent of this node is \emph{not} a vertical stack, the normalized form is rearranged into a horizontal stack of vertical stacks. This means that the normalized form of \inlinesign{za} is always \code{[\{vv\}\{vv\}]}, never \code{\{[vv][vv]\}}.
	\item Conversely, if a horizontal stack contains only vertical stacks, and all those vertical stacks have the same number of elements, and the parent of this node \emph{is} a vertical stack, the normalized form is rearranged into a vertical stack of horizontal stacks. This reduces the total number of stacks, since a vertical stack inside a vertical stack is removed by the third special case.
	\item If a vertical stack contains one or more horizontal stacks, and any of those stacks contains a horizontal stroke and a Winkelhaken at the right or left end, remove those Winkelhakens from the ends and rearrange them into vertical stacks of their own. Then combine those vertical stacks together as a horizontal stack: \code{\{[hc]h\}} becomes \code{[\{\}\{hh\}\{c\}]} (removing one Winkelhaken from the right and none from the left). The empty and singleton stacks are then simplified by the other cases above. This handles the ambiguity of signs like \htt{u\shin{}}; the difference between \inlinesign{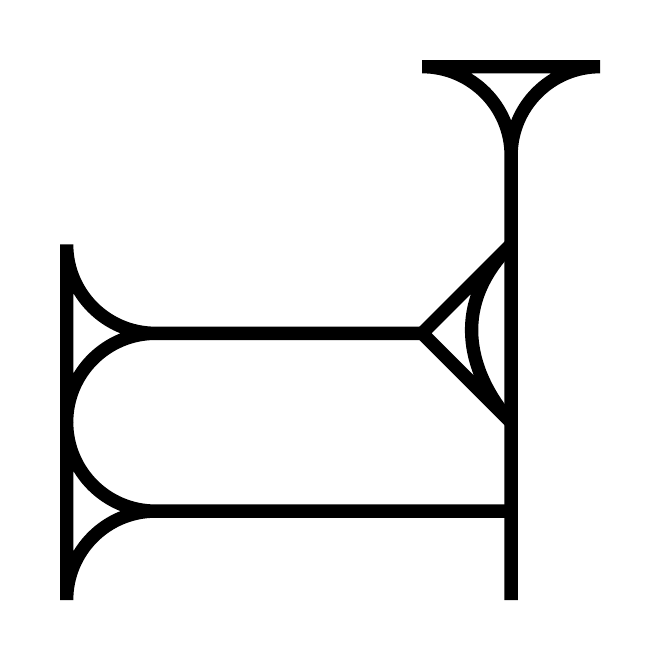} and \inlinesign{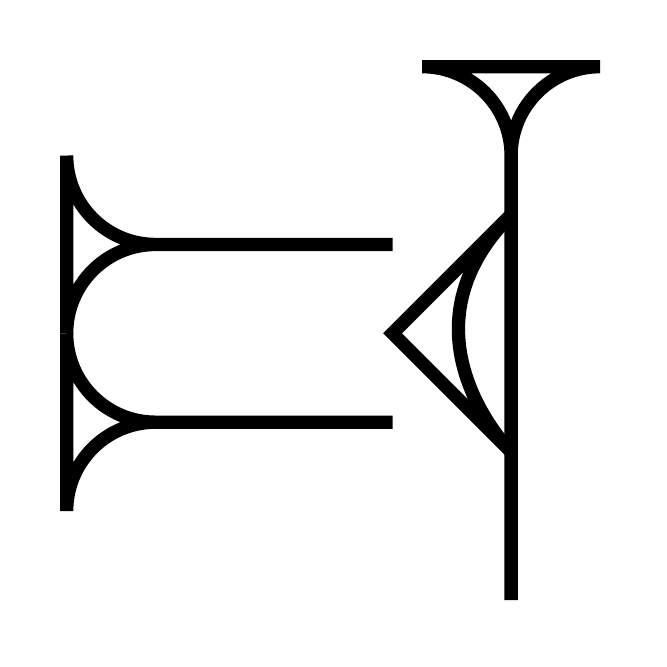} is clear in the trees, but not at all clear on the actual clay, and this ensures that the normalized form always has the Winkelhaken outside the vertical stack.
	\item Define a \emph{mixed} superposition to be a superposition indirectly containing both horizontal and vertical strokes. If a vertical stack contains only horizontal elements and a mixed superposition, or a horizontal stack contains only vertical elements and a mixed superposition, the normalized form puts the stack inside the superposition. This is the strangest normalization rule, and its purpose is to ensure that patterns like \code{\{hh(hv)\}}, \code{\{h(\{hh\}v)\}}, and \code{(\{hhh\}v)} have the same normalized form, since they are nearly impossible to distinguish on actual clay.
	\item If none of these special cases apply, the normalized form of the composition is the composition of the normalized forms of its children.
\end{itemize}
\end{itemize}

The end result is that aesthetic variations, like the lengths of strokes or the size of the gaps between them, will not affect searching or comparisons.

\subsection{Encompassing}
\label{subsec:encompass}

\begin{wrapfigure}[14]{L}{0.23\textwidth}
\centering
\includegraphics[width=0.18\textwidth]{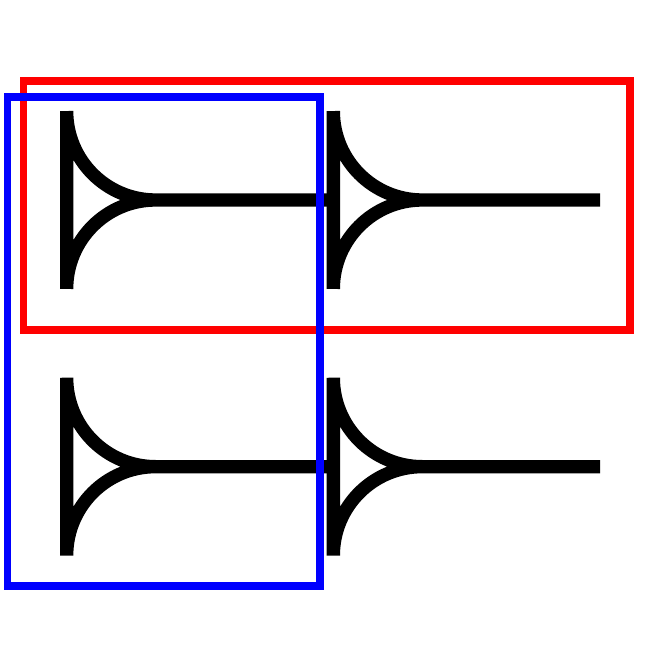}
\caption{An illustration of why subtree matching is not sufficient for sign identification.}
\label{fig:hhhh}
\end{wrapfigure}

The most crucial algorithm is termed \emph{encompassing}. For the search system to be effective, a tree must ``encompass'' any subset of its strokes, as long as the relationships between those strokes are preserved; the strokes should not have to be contiguous, and the relationships to any other strokes not in the subset should not matter. In other words, a sign should ``encompass'' any part a student might search for, even if other parts of the sign are damaged or difficult to read.

The algorithm given here is significantly more complex than previous approaches to recursive searching. \citet{downes}, for instance, uses standard substring matching in Excel, and similar algorithms exist to match subtrees\footnote{What linguists would term `constituents' in syntax: one particular node and everything dominated by it.}. \citet{paleocodage} likewise uses standard metrics of string similarity. However, subtree matching is not sufficient for our purposes. Consider the sign in figure~\ref{fig:hhhh}. It's impossible for both the red component and the blue component to be subtrees. But both are perfectly reasonable ways for a student to search for the sign! Something better is needed.

The encompassing algorithm is the most crucial part of this system. This relation allows students to search for whichever part of a sign is clearest and most readable. They aren't limited to the leftmost part of the sign, or even a complete subtree; \emph{any} visible strokes can be used to narrow the search. (See section~\ref{subsec:exsearch} for an example of this.)

The algorithm for this is, like the others, recursive.
\begin{itemize}
\item A stroke encompasses a stroke of the same type.
\item Any stroke encompasses a wildcard.
\item A composition \(A\) encompasses a node \(B\) if any child of \(A\) encompasses \(B\).
\item A superposition \(A\) encompasses a superposition \(B\) if every child of \(B\) is encompassed by some child of \(A\)\footnote{Notably, this step does not check that those children of \(A\) are distinct. This makes the implementation much simpler, but can lead to the occasional false positives, such as \code{([vh]c)} encompassing \code{(vh)}.}.
\item A stack \(A\) encompasses a stack \(B\) if:
\begin{itemize}
	\item \(A\) and \(B\) have the same type (vertical or horizontal), \textbf{and}
	\item Every child of \(B\) is encompassed by some child of \(A\), \textbf{and}
	\item For any children of \(B\) \(x\) and \(y\), if \(x\) precedes \(y\), then the child of \(A\) that encompasses \(y\) does not precede the child of \(A\) that encompasses \(x\). This ensures that \code{[vh]} does not encompass \code{[hv]}, but \code{[hcv]} does. In other words, it loosely enforces an ordering.
\end{itemize}
\end{itemize}

A slight modification of this algorithm can also return a list of the strokes checked in the first bullet point. In the interface, this is used to highlight the matching part of each search result, displaying those strokes in a different color (see figures~\ref{fig:exance} and \ref{fig:exdug}).

\subsection{Interface}
\label{subsec:interface}

Finally, some sort of interface is needed for students to actually make use of these algorithms. The prototype consists of a \emph{canvas} and a \emph{search engine}, available at \url{https://dstelzer.pythonanywhere.com/canvas.html}.

The canvas is designed to help students input recursive codes. It has a text box to enter a code, and displays the graphical result next to it, updating in real time. The position of the cursor in the text box is reflected with a green bar in the output, showing where a newly-typed stroke would appear, and any strokes highlighted in the text box are colored green. This is intended to help users develop an intuition for the \stelzer{} system (see figure~\ref{fig:ui1}).

The search engine then takes a pattern entered through the canvas and displays a list of signs that encompass that pattern, using the algorithm from section~\ref{subsec:encompass} (see figure~\ref{fig:ui2}). The database behind this currently contains all signs and major variants from \citet{hzl}, and the results can be sorted by \citeauthor{hzl}'s index number, sign usage (phonogram, heterogram, logogram, semagram), or number of wedges\footnote{Specifically, sign ``complexity'' is defined as as the number of leaves in the normalized form of the most standard or canonical shape of the sign---that is, Gottstein's ``category'' number---and users can sort the results by this value. Since different variants might have different numbers of strokes, this value is defined for each sign, not for each variant.}.

\section{Examples}
\label{sec:examples}

\subsection{Encoding}

\begin{figure}[h]
\centering
\Large
{\huge \inlinesign{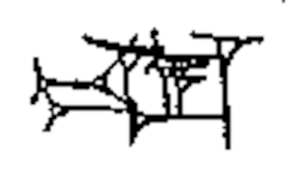} \inlinesign{nag}} (\sign{ka\x{}a}) = \code{W[\{0,[hc],h\},v,\{h,[v!v2!]M,h\},v]}
\begin{forest}
[\code{W}
	[\inlinesign{hstack}
		[\inlinesign{vstack}
			[\code{0}]
			[\inlinesign{hstack}
				[\inlinesign{h}]
				[\inlinesign{c}]
			]
			[\inlinesign{h}]
		]
		[\inlinesign{v}]
		[\inlinesign{vstack}
			[\inlinesign{h}]
			[\code{M}
				[\inlinesign{hstack}
					[{\inlinesign{v}\code{!}}]
					[{\inlinesign{v}\code{2!}}]
				]
			]
			[\inlinesign{h}]
		]
		[\inlinesign{v}]
	]
]
\end{forest}
\vspace{1em}
\caption{A more complex recursive encoding: the sign \sign{KA\x{}A} (logographic \smr{nag} `drink'), rendered in color for figure~\ref{fig:nag}. Node adjustments like \code{M} act as their own non-branching nodes in the tree, while stroke modifiers like \code{2} are parsed as part of a leaf. The commas are ignored by the parser but can be used to clarify the boundaries between sub-units.}
\label{fig:complex}
\end{figure}

\begin{figure}[h]
\centering
\large
{\Huge \inlinesign{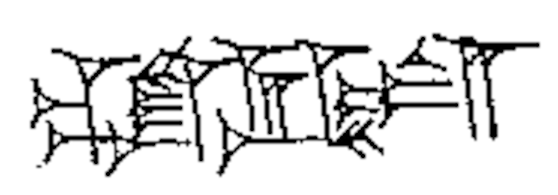} \inlinesign{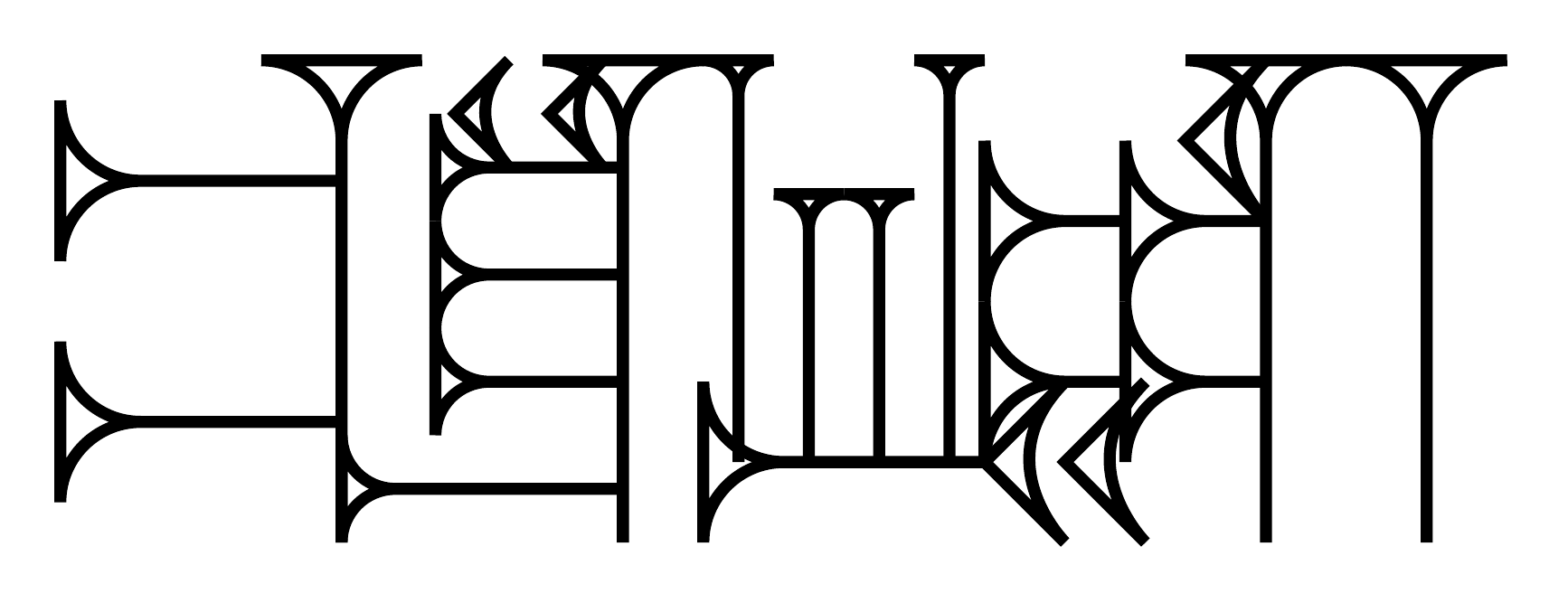}} (\sign{umbin}) = \code{X[ \{hh\}v, \{[0'cc]h'h'h'h\}v, 0"', \{[vv'v'v]h\}, \{[0c],[hh],[hh],[cc0]\}, vv ]}

\begin{forest}
for tree={l sep+=1em,s sep-=2ex}
[\code{X}
	[\inlinesign{hstack}
		[\inlinesign{vstack}
			[\inlinesign{h}]
			[\inlinesign{h}]
		]
		[\inlinesign{v}]
		[\inlinesign{vstack}
			[\inlinesign{hstack}
				[\code{0'}]
				[\inlinesign{c}]
				[\inlinesign{c}]
			]
			[\inlinesign{h}\code{'}]
			[\inlinesign{h}\code{'}]
			[\inlinesign{h}\code{'}]
			[\inlinesign{h}]
		]
		[\inlinesign{v}]
		[\code{0'"}]
		[\inlinesign{vstack}
			[\inlinesign{hstack}
				[\inlinesign{v}]
				[\inlinesign{v}\code{'}]
				[\inlinesign{v}\code{'}]
				[\inlinesign{v}]
			]
			[\inlinesign{h}]
		]
		[\inlinesign{vstack}
			[\inlinesign{hstack}
				[\code{0}]
				[\inlinesign{c}]
			]
			[\inlinesign{hstack}
				[\inlinesign{h}]
				[\inlinesign{h}]
			]
			[\inlinesign{hstack}
				[\inlinesign{h}]
				[\inlinesign{h}]
			]
			[\inlinesign{hstack}
				[\inlinesign{c}]
				[\inlinesign{c}]
				[\code{0}]
			]
		]
		[\inlinesign{v}]
		[\inlinesign{v}]
	]
]
\end{forest}
\vspace{1em}
\caption{\smr{umbin} `fingernail', the most complex sign in the \emph{Zeichenlexikon} by number of strokes. It can be built recursively out of a series of vertical components stacked horizontally.}
\label{fig:morecomplex}
\end{figure}

More complex examples of the recursive encoding can be seen in figures~\ref{fig:complex} and \ref{fig:morecomplex}.

\subsection{Searching}
\label{subsec:exsearch}

\begin{figure}[h]
\centering
\includegraphics[width=0.25\linewidth]{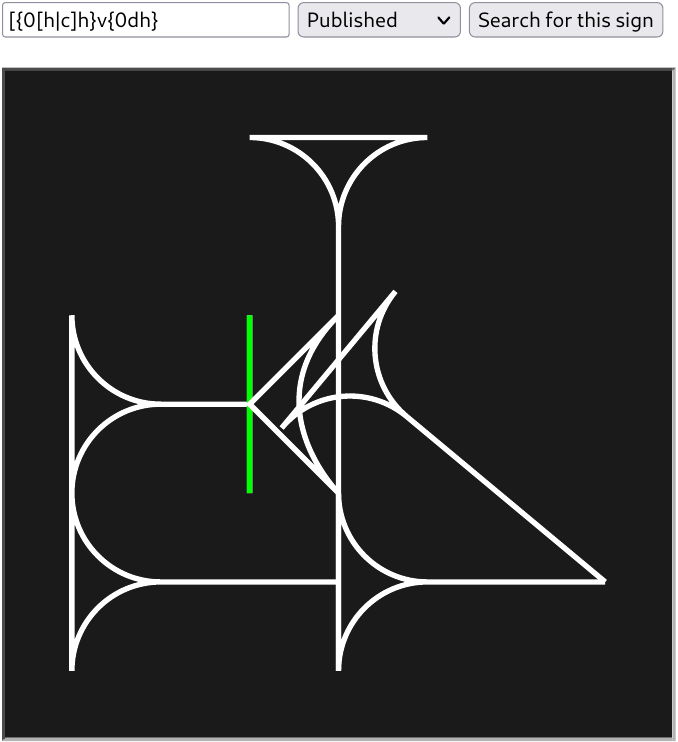}
\caption{The ``canvas'' part of the interface, helping users enter \stelzer{} codes by showing the result in real time. The code entered here doesn't need to be complete (note the \code{[} opened at the left but never closed), and the position of the cursor is marked with a green bar (between the horizontal and Winkelhaken). This indicates where in the sign a newly-typed stroke will be inserted. The drop-down at the top changes the rendering style, letting users customize it to their preference.}
\label{fig:ui1}
\end{figure}

\begin{figure}[h]
\includegraphics[width=\linewidth]{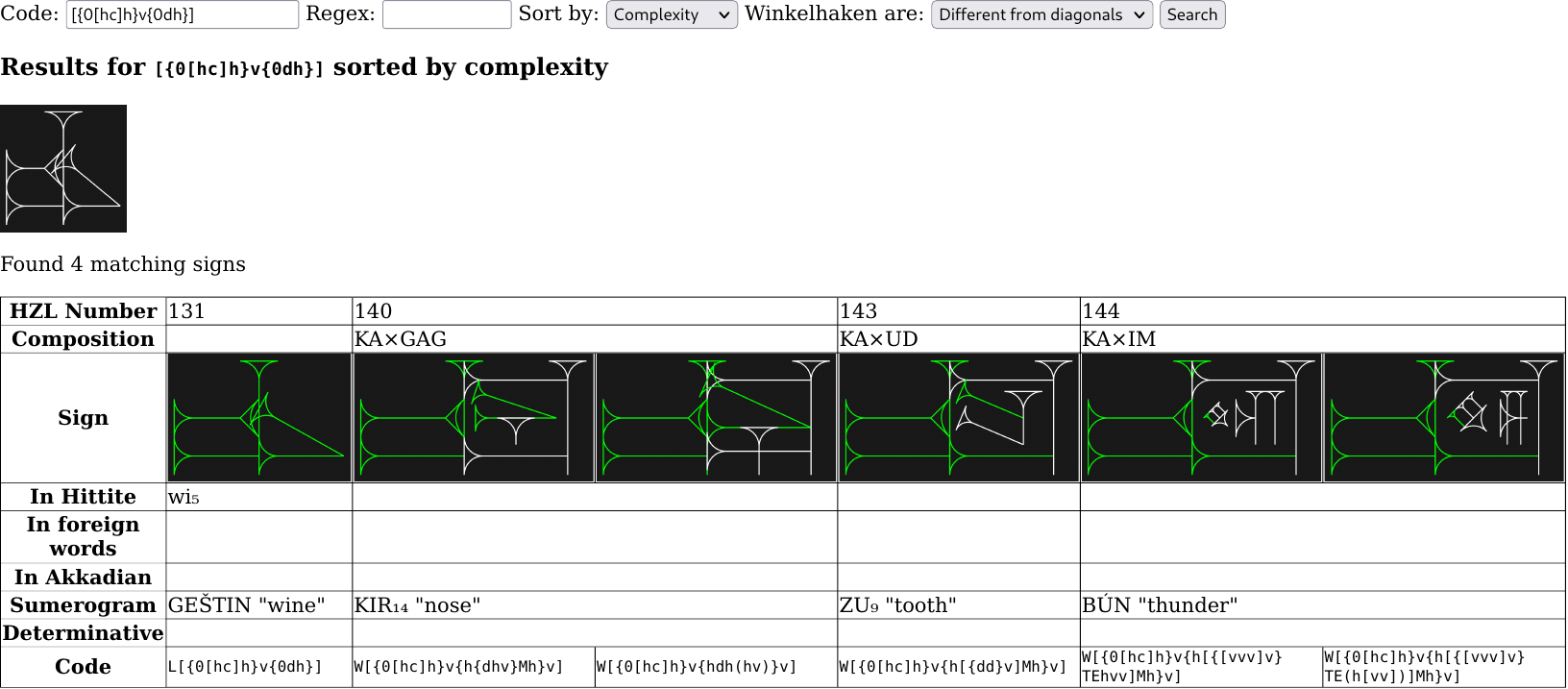}
\caption{The ``search'' part of the interface, showing all the signs encompassing the code the user created with the canvas. The matching strokes are highlighted in green. Other options allow the user to narrow down signs by name and change the sorting method or normalization mode.}
\label{fig:ui2}
\end{figure}

\begin{figure}[h]
\includegraphics[height=1.5in]{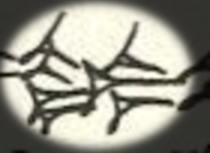} \hfill
\includegraphics[height=1.5in]{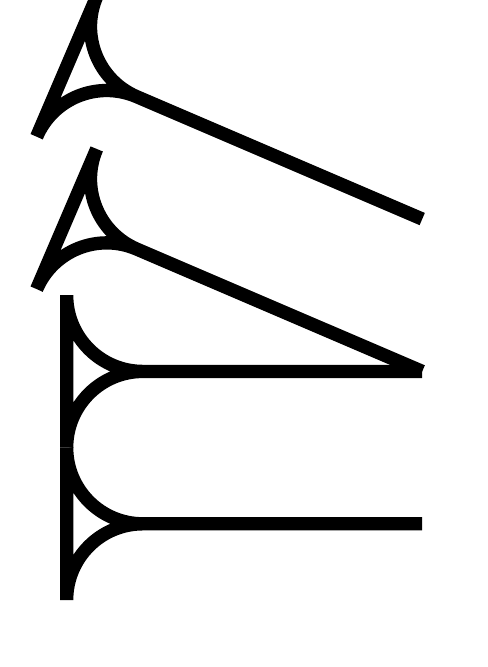} \hfill
\includegraphics[height=1.5in]{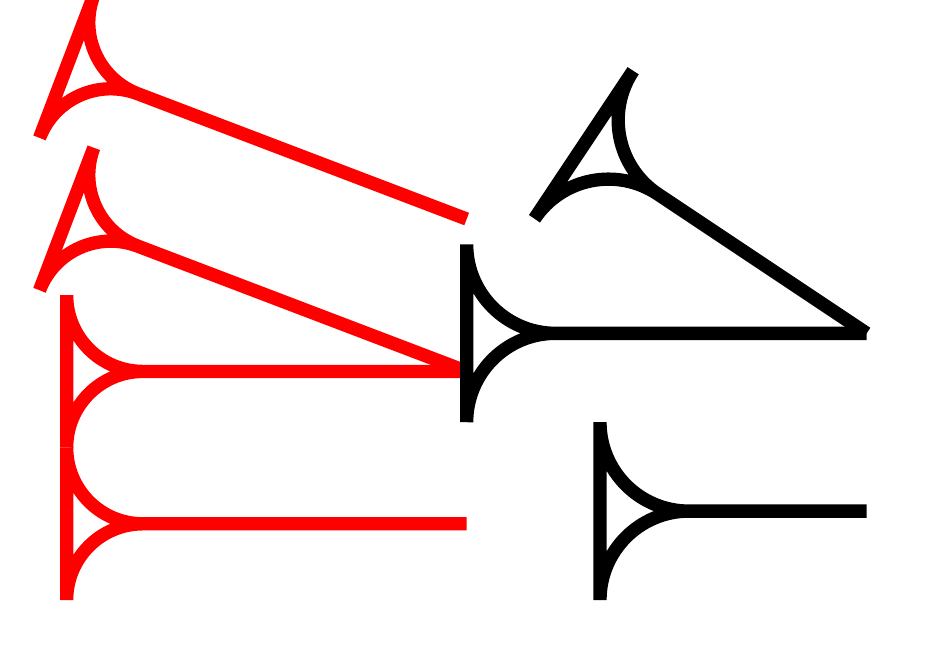}
\caption{The process of searching for a complete sign: \smr{an\shin{}e} `donkey'. The user looks for the part of the sign that seems easiest to encode, in this case, the four strokes on the left: \code{\{ddhh\}}. The program then shows them a list of all six signs which encompass this component, and the user can identify the right sign from among them.}
\label{fig:exance}
\end{figure}

\begin{figure}[h]
\includegraphics[height=1.5in]{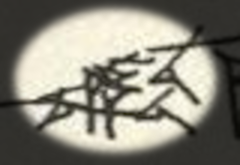} \hfill
\includegraphics[height=1.5in]{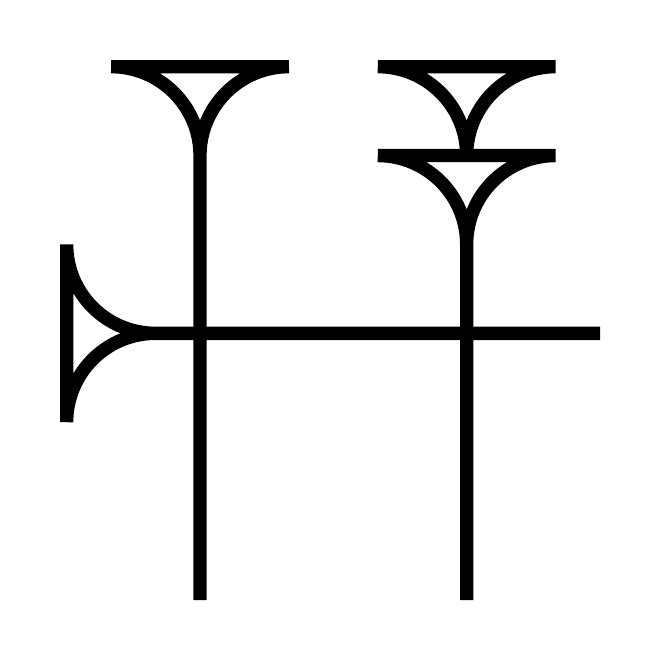} \hfill
\includegraphics[height=1.5in]{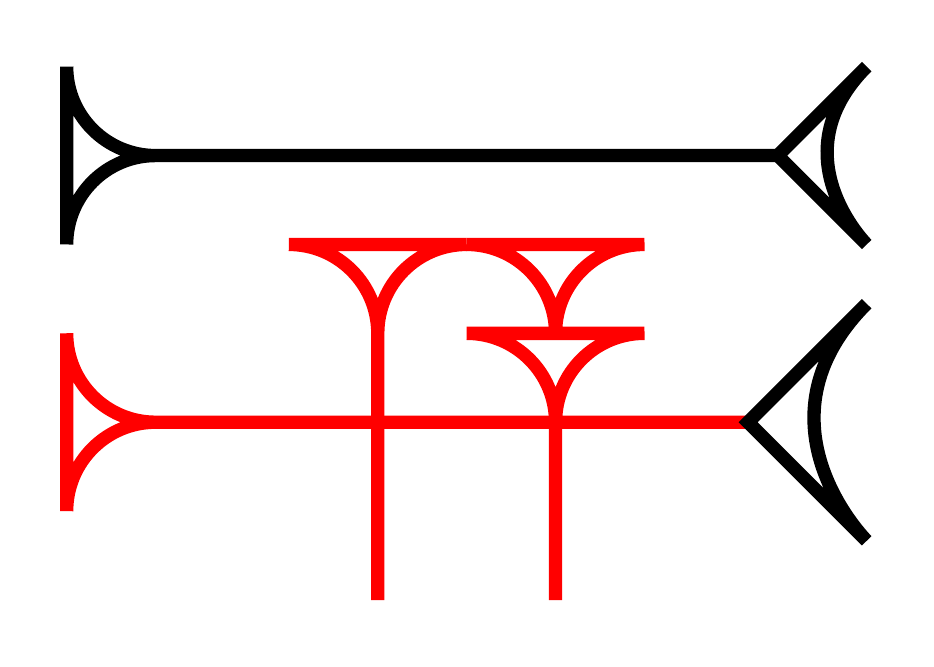}
\caption{The process of searching for a broken sign: \smr{dug} `vessel'. Since the left side of this sign is broken, the user encodes the bottom part: the horizontal stroke intersecting an \sign{A} sign, \code{(h[vv2])}. The program can again produce a list of all signs encompassing this element (seven, in this case), allowing the user to identify the right one.}
\label{fig:exdug}
\end{figure}

Examples of the search process can be seen in figures~\ref{fig:exance} and \ref{fig:exdug}. The interface for the search can be seen in figures~\ref{fig:ui1} and \ref{fig:ui2}.

\section{Conclusion}

This paper proposes a new method of encoding the shape of a cuneiform sign, as a tree made up of strokes and compositions. As a result, computers can now work with the shape of a sign in a way that wasn't possible before. At its most basic level, this can be used for rendering, turning this recursive encoding into a picture of the sign (as in figure~\ref{fig:font}).

But rendering is only the tip of the iceberg. Using the interface discussed in section~\ref{subsec:interface}, students can now search for a sign based on any strokes that are visible, regardless of damage or obscurity. It can also serve as an intermediate form for machine learning approaches to cuneiform, allowing problems to be factored into smaller steps than were possible before. I believe these recursive approaches have potential far beyond what's been tested so far, and may be a significant step forward in computational analysis of cuneiform writing.

The code discussed in this article can be found at \url{https://bitbucket.org/dstelzer/hantatallas}, and can be used at \url{https://dstelzer.pythonanywhere.com/canvas.html}.

\subsection{Acknowledgements}

I would like to thank my thesis advisor, Ryan Shosted, who introduced me to cuneiform studies in the first place and funded my experiments with this system. I'd also like to thank the user Yellow Sky on StackExchange, who spurred all of this by directing my attention to Downes' work back in 2020.

\printbibliography

\end{document}